\documentclass{article}

\usepackage[preprint,nonatbib]{neurips_2025}

\usepackage[utf8]{inputenc} 
\usepackage[T1]{fontenc}    
\usepackage{url}            
\usepackage{booktabs}       
\usepackage{amsfonts}       
\usepackage{nicefrac}       
\usepackage{microtype}      
\usepackage{xcolor}
\usepackage{graphicx}
\usepackage{amsmath}
\usepackage{multirow}
\usepackage{multicol}
\usepackage{colortbl}
\usepackage{tcolorbox}
\usepackage{subcaption}
\usepackage{pifont}
\usepackage{listings}
\usepackage{xr}
\usepackage{wrapfig}

\newcommand{\VarSty}[1]{\textcolor{blue}{\texttt{#1}}}
\usepackage[backend=biber, style=numeric]{biblatex} 
\usepackage{hyperref}
\definecolor{light-gray}{gray}{0.6}
\definecolor{front-color}{HTML}{F5FFFA}
\definecolor{Gray}{gray}{0.93}
 
\graphicspath{{./fig}}

\title{SynPO: Synergizing Descriptiveness and Preference Optimization for Video Detailed Captioning}

\makeatletter
\renewcommand{\@fnsymbol}[1]{\ifcase#1\or *\or \dagger\fi}
\makeatother

\author{
  Jisheng Dang$^{1,2,5}$\thanks{Equal contribution.}~~
  Hao Ye$^{2}$\footnotemark[1]~~
  Yizhou Zhang$^{2}$\footnotemark[1]~~
  Teng Wang$^{3}$\thanks{Corresponding author.}~~
  Siming Chen$^{2}$\\
  \textbf{Huicheng Zheng$^{1}$~~Yulan Guo$^{1}$~~Jianhuang Lai$^{1}$~~Bin Hu$^{4}$}\\
  $^{1}$ Sun Yat-Sen University \ 
  $^{2}$ Lanzhou University \ 
  $^{3}$ The University of Hong Kong \\
  $^{4}$ Beijing Institute of Technology \ 
  $^{5}$ National University of Singapore
}

\begin{document}

\maketitle
\vspace{-5mm}
\begin{abstract}
Fine-grained video captioning aims to generate detailed, temporally coherent descriptions of video content. However, existing methods struggle to capture subtle video dynamics and rich detailed information. In this paper, we leverage preference learning to enhance the performance of vision-language models in fine-grained video captioning, while mitigating several limitations inherent to direct preference optimization (DPO). First, we propose a pipeline for constructing preference pairs that leverages the intrinsic properties of VLMs along with partial assistance from large language models, achieving an optimal balance between cost and data quality. 
Second, we propose Synergistic Preference Optimization (SynPO), a novel optimization method offering significant advantages over DPO and its variants. 
SynPO prevents negative preferences from dominating the optimization, explicitly preserves the model's language capability to avoid deviation of the optimization objective, and improves training efficiency by eliminating the need for the reference model. 
We extensively evaluate SynPO not only on video captioning benchmarks (e.g., VDC, VDD, VATEX) but also across well-established NLP tasks, including general language understanding and preference evaluation, using diverse pretrained models. 
Results demonstrate that SynPO consistently outperforms DPO variants while achieving 20\% improvement in training efficiency.
Code is available at \href{https://github.com/longmalongma/SynPO}{https://github.com/longmalongma/SynPO}.
\end{abstract}
\section{Introduction}

\label{sec:intro}
Fine-grained video captioning aims to generate detailed and coherent textual descriptions that precisely capture video contents. This task necessitates the recognition of salient actions and objects, while also modeling fine-grained visual features and temporal dynamics. Recent studies~\cite{cheng2024videollama2advancingspatialtemporal,jin2024chat,cai2024matryoshka,li2024llama} have primarily employed Vision-Language Models (VLMs) for video captioning. These methods~\cite{li2023blip,li2023videochat,zhang2023video} typically utilize pre-trained vision encoders and Large Language Models (LLMs), with a connector module linking them. By training on video-text pairs, these models aim to align visual and textual representations, thereby enhancing their ability to understand and describe video content~\cite{wang2023chatvideo} effectively.

\begin{figure}[t]
    \centering
    \vspace{-4mm}
    \includegraphics[width=\linewidth]{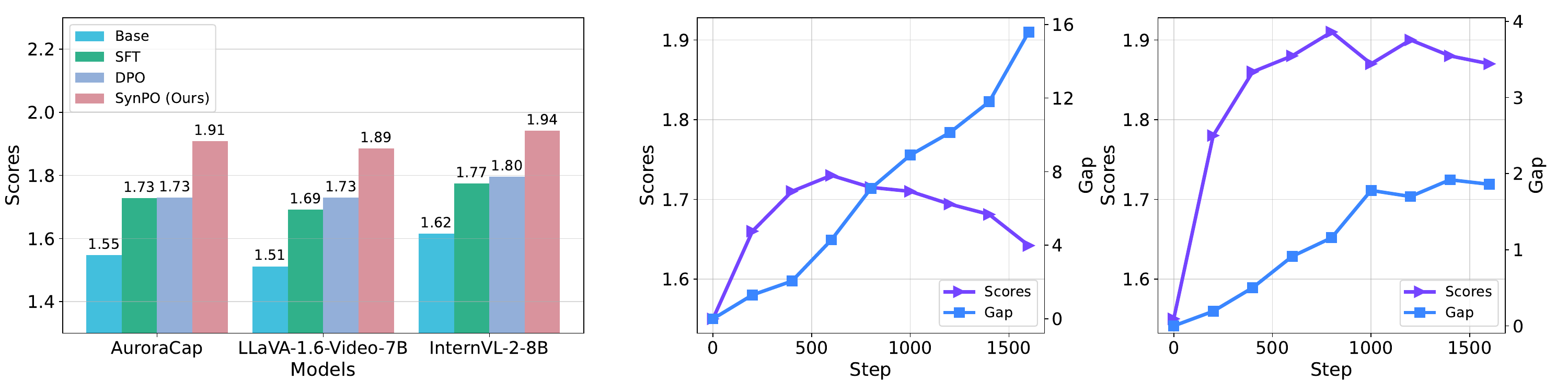}
    \caption{\textbf{Left:} SynPO significantly outperforms other methods in different models on VDC benchmark~\cite{chai2024auroracap}.
    \textbf{Middle:} Language capability degradation occurs during the latter training stages in DPO. Training collapses and is biased towards maximazing positive-negative reward gap.
    \textbf{Right:}  SynPO mitigates degradtion successfully and resolves the issue of optimization objectives shifting from language capability to ranking differentiation. Its performance significantly outperforms that of DPO.
    }
    \label{fig:fig1}
    \vspace{-6.5mm}
\end{figure}

Direct Preference Optimization (DPO)~\cite{rafailov2024direct} is a fine-tuning method that aligns models with stipulated preferences using high-quality preference pairs. Recent work has successfully adapted DPO and its variants to video understanding tasks, significantly improving model performance~\cite{li2025templetemporalpreferencelearningvideo,li2025temporalpreferenceoptimizationlongform}. Thus, integrating DPO into fine-grained video captioning to enhance the model’s ability to capture temporal dynamics and detailed descriptions present a promising direction. However, two critical challenges currently degrade its performance in fine-grained video captioning:
(1) the scarcity of high-quality video-text alignment pairs, which are essential for preference learning; (2) DPO typically suffers from the simultaneous decrease in both positive and negative reward values~\cite{pal2024smaugfixingfailuremodes, chen2024improvedpreferenceoptimizationpipeline}, leading to a potential objective optimization deviation from focusing on generation quality to merely discriminating between preferences~\cite{feng2024analyzingunderstandinglimitationsdpo}, as shown in Figure~\ref{fig:fig1} (middle).

Recently, the  proposed VDC benchmark~\cite{chai2024auroracap} is well-curated, but limited in scale and partially reliant on manual annotations. Other video captioning datasets, such as MSRVTT~\cite{xu2016msr}, VATEX~\cite{wang2020vatexlargescalehighqualitymultilingual}, MSVC~\cite{cheng2024videollama2advancingspatialtemporal}, etc., typically provide overly brief captions, falling short in fine-grained video captioning. Besides, these datasets lack preference pairs, and thus cannot be prepared for DPO. To construct preference pairs, many existing methods~\cite{an2025agfsyncleveragingaigeneratedfeedback,Li_2024} rely on a stronger VLM to score multiple outputs from the same prompt. While straightforward, this approach is impractical: small teams face prohibitive API costs, while developers of powerful models often lack access to a stronger scoring model. Some studies attempt to circumvent this limitation by generating negative preferences, such as through atypical item substitution~\cite{xie2024vdpomitigatinghallucinationlarge} or temporal perturbation~\cite{li2025templetemporalpreferencelearningvideo}. However, these methods primarily focus on negative samples and fail to produce higher-quality positive preferences.

We present an automated pipeline for constructing high-quality preference pairs for fine-grained video captioning. Given the same input, we generate multiple alternative outputs using a VLM. These candidates are then scored leveraging intrinsic properties of the VLM itself, such as its self-consistency~\cite{10298561} and the enhanced ability to capture details in short videos, with limited assistance from an LLM. The top and bottom scores are selected as positive and negative preferences, respectively, forming our constructed preference dataset. Compared to existing approaches, our method achieves an optimal balance between cost efficiency and high-quality preference pair construction.

We propose an improved optimization method for DPO, termed Synergistic Preference Optimization (SynPO). Our SynPO features three critical advantages: (1) It reformulates the reward gap computation to prevent the influence of negative preferences from dominating the optimization process, thereby fundamentally addressing the issue of simultaneous decreases in both positive and negative reward values; (2) It introduces an additional reward term in the loss function that explicitly encourages language capability, helping to maintain the model's generative performance and prevent objective drift during optimization, shown in Figure \ref{fig:fig1} (right); (3) It eliminates the need for a reference model during training, resulting in an approximately 20\% improvement in training efficiency.

Our experiments across multiple models and datasets demonstrate that our data construction pipeline can generate high-quality preference datasets with considerable generality. In addition to video captioning benchmarks (e.g., VDC~\cite{chai2024auroracap}, VDD~\cite{li2023videochat}, VATEX~\cite{wang2020vatexlargescalehighqualitymultilingual}, MSR-VTT~\cite{xu2016msr}), we conduct comparisons of SynPO with various DPO variants~\cite{Ethayarajh2024KTOMA, xu2024contrastive, pal2024smaugfixingfailuremodes, Azar2023AGT, meng2024simposimplepreferenceoptimization} across multiple NLP tasks, which include preference evaluation tasks (e.g., MT-Bench~\cite{zheng2023judging}, AlpacaEval2~\cite{li2023alpacaeval}) and downstream applications (e.g., tasks from the Huggingface Open LLM Leaderboard~\cite{open-llm-leaderboard-v1,open-llm-leaderboard-v2}). As shown in Figure \ref{fig:fig1} (left), extensive results indicate that SynPO significantly outperforms DPO and its variants.

The contributions of this paper are three-fold: 1) We propose a novel pipeline that automatically generates high-quality preference pairs for fine-grained video captioning by leveraging a VLM's intrinsic self-consistency and detail-capturing ability; 2) We introduce SynPO, an improved DPO method that prevents deviations during optimization via reformulated reward computation and incorporates an explicit language reward to maintain generation quality; 3) Extensive experiments on video captioning demonstrate SynPO's superiority over six DPO variants. Our approach also achieves superior results on NLP preference tasks and Open LLM Leaderboard, verifying its effectiveness across domains.

\section{Related Work}
\label{sec:rel}

\textbf{Video Captioning.}
Early video captioning methods employed template-based approaches~\cite{guadarrama2013youtube2text} or RNN-based encoder-decoder frameworks~\cite{venugopalan2014translating}, but were limited in modeling long-range dependencies. Transformers~\cite{vaswani2017attention}~\cite{dao2022flashattentionfastmemoryefficientexact}~\cite{jin2024mohmultiheadattentionmixtureofhead}~\cite{dai2019transformerxlattentivelanguagemodels}~\cite{parmar2018imagetransformer} and vision-language pre-training significantly advanced the field, with models like CLIP4Caption~\cite{tang2021clip4caption} and SwinBERT~\cite{lin2022swinbert} improving video-text alignment. Recent VLMs~\cite{cheng2024videollama2advancingspatialtemporal, li2024llama} adapt multimodal architectures such as BLIP-2~\cite{li2023blip} and LLaVA~\cite{liu2023visual} to video, though many struggle with temporal dynamics. Newer approaches like VideoLLaMA~\cite{zhang2023video} and ChatVideo~\cite{wang2023chatvideo} better model sequential structure for enhanced comprehension.

\textbf{Reinforcement Learning and Preference Learning.}
Reinforcement Learning from Human Feedback (RLHF)~\cite{christiano2017deep} aligns LLMs with human preferences~\cite{kaufmann2024surveyreinforcementlearninghuman,Ouyang2022TrainingLM,stiennon2020learning} through supervised fine-tuning~\cite{zhou2024lima}, reward modeling~\cite{gao2023scaling}, and policy optimization, improving instruction-following~\cite{ouyang2022training} and safety~\cite{bai2022training}. To simplify RLHF's complexity, offline methods like DPO~\cite{rafailov2024direct} bypass explicit reward modeling, inspiring variants such as IPO~\cite{azar2024general}, ORPO~\cite{hong2024orpo} and others~\cite{yuan2024rrhf, xucontrastive, xiao2024leverage,wu2024alphadpoadaptiverewardmargin}.

\begin{figure}[t]
    \centering
    \includegraphics[width=\textwidth]{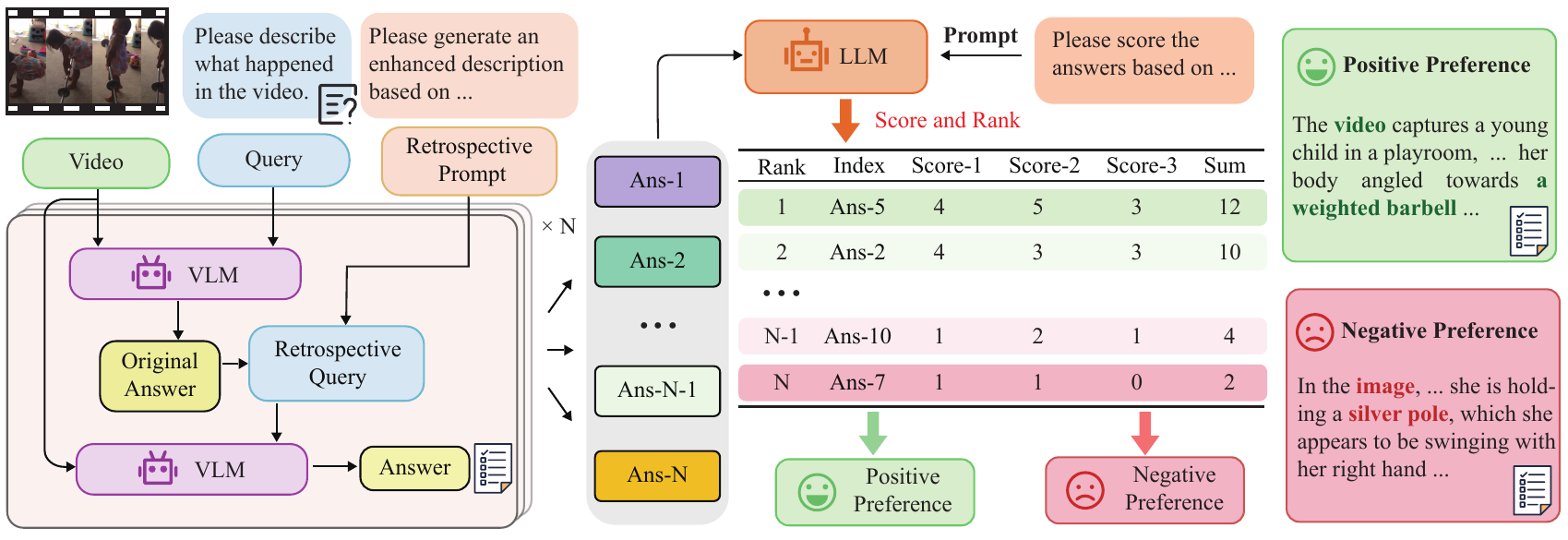}
    \caption{Overview of dataset construction pipeline. A VLM first generates multiple candidate captions for each video with the self-retrospective strategy. Then the candidate captions are scored by an LLM based on three criteria (i.e., factuality, linguistic fluency, and self-consistency) to select positive and negative preferences.}
    \label{fig:workflow}
\end{figure}

\section{Constructing Long Video Caption Preference Pairs}\label{sec:data construction}
\subsection{Enhanced Model Inference}
To address the challenges of hallucination and insufficient detail generation in fine-grained video captioning, we incorporate contrastive decoding and a self-retrospective strategy. These methods target complementary aspects: contrastive decoding reduces hallucinations and enhances precision, while the self-retrospective strategy encourages the model to capture more detailed information.

Contrastive decoding was initially proposed by \cite{leng2023mitigatingobjecthallucinationslarge} to reduce object hallucinations and subsequently improved by \cite{zhang2025eventhallusiondiagnosingeventhallucinations}, who introduced a more efficient variant that contrasts logits from sparse frame samples with those from full sequences. In this work, the improved contrastive decoding method is adopted to suppress overconfidence in noisy or irrelevant features, thereby improving factual consistency and detail accuracy in generated captions. 

The self-retrospective strategy, proposed by \cite{ahn2025isrdpoaligninglargemultimodal}, operates outside the decoding process and enhances comprehension through iterative refinement. Feeding the model's own outputs back into its input enables a form of retrospective reasoning~\cite{madaan2023selfrefineiterativerefinementselffeedback}, allowing predictions to be refined in light of prior generations. We adapt this method to video captioning by using the initial caption as contextual input for subsequent refinement steps, enabling richer and more coherent descriptions.

These two strategies are integrated by applying contrastive decoding at both stages of the self-retrospective process. Specifically, contrastive decoding is utilized when generating the initial caption and again during the refinement step. This ensures that each iteration benefits from a reduction in hallucinations and an improvement in fidelity, while also leveraging the iterative refinement capabilities of the self-retrospective strategy to enhance descriptive richness and linguistic fluency.

\subsection{Dataset Construction Pipeline}
We propose an automated pipeline for constructing high-quality preference pairs specifically designed for fine-grained video captioning. The overall framework is designed to address the limitations of existing datasets, such as limited scale, insufficient detail in captions, and the lack of human-like preference annotations~\cite{chai2024auroracap,wang2020vatexlargescalehighqualitymultilingual,cheng2024videollama2advancingspatialtemporal}. Our method leverages both the intrinsic properties of VLMs and the reasoning capabilities of LLMs to generate diverse and reliable preference pairs without dependence on costly multimodal scorers.

The core strategy involves generating multiple candidate captions per video using a single VLM under the same prompt. These candidates are then scored using a novel three criterion evaluation framework that combines factuality, modality correctness, and self-consistency~\cite{10298561}. Based on the aggregated scores across all three criteria, the captions with the highest and lowest total scores are selected as positive and negative preferences, respectively, forming our final dataset. In the following, we describe each scoring criterion in detail (full prompts provided in Appendix \ref{sec:full prompts}).

\textbf{Criterion 1: Factuality through Temporal Decomposition.}
Due to input length limits in most VLMs, processing long videos directly often causes detail loss and hallucinations~\cite{li2025templetemporalpreferencelearningvideo}. To mitigate this, we divide each video into short clips, process them independently with the VLM to generate clip-level captions, and concatenate these into a reference set. An LLM then assesses the consistency between the full-video caption and the reference set, focusing particularly on factual alignment. This approach enhances detail preservation and mitigates hallucinations. Scores range from 0 to 5.

\textbf{Criterion 2: Instruction Fidelity, Linguistic Fluency and Objectivity.}
Video captions generated by the VLM are assessed by an LLM according to the following criteria:
(1) Instruction fidelity: Whether the caption meets the requirements of the corresponding prompt;
(2) Linguistic fluency: Whether the description is natural and coherent, using language appropriate for describing a video (e.g., avoid calling a video an "image") ;
(3) Objectivity: Minimizing subjective or illogical content.
Each caption receives an overall score between 0 and 5, ensuring semantically accurate and linguistically well-formed outputs.

\textbf{Criterion 3: Self-consistency through Multi-sample Analysis.}
Inspired by self-consistency methods in NLP~\cite{rueda2025understandingllmscientificreasoning,liu2025enhancingmathematicalreasoninglarge,sanzguerrero2025correctiveincontextlearningevaluating}, we apply it to video captioning by assessing the stability of key entities, actions, and temporal dynamics across multiple generations. Specifically, the VLM generates 
$n$ diverse captions via high-temperature sampling. An LLM analyzes their similarity, rewarding consistent patterns and penalizing outliers through a majority voting mechanism. Given its narrower discriminative capacity compared to the other two criteria, this metric uses a 0 to 3 scoring range.

\section{SynPO: Synergistic Preference Optimization}
\label{sec:synpo}

\subsection{Preliminary}
RLHF is a methodology that leverages human evaluations to optimize models through reinforcement learning paradigms. The core workflow of RLHF typically consists of two main stages: First, a reward model is trained using human feedback data as:
\begin{equation}\label{eq:reward_model}
    \mathcal{L}_R(r_{\phi}, \mathcal{D}) = -\mathbb{E}_{(x, y_w, y_l)\sim \mathcal{D}}\bigl[\log \sigma(r_{\phi}(x, y_w)- r_{\phi}(x, y_l))\bigr],
\end{equation}
where $ r_{\phi} $ denotes the reward function parameterized by $ \phi $, 
$ (x, y_w, y_l) $ represents a triplet consisting of input prompt $ x $ from the preference dataset $\mathcal{D}$, 
$y_w$ and $y_l$ denote the positive and negative preferences respectively,
and $ \sigma(\cdot) $ is the logistic sigmoid function.

Second, the learned reward model is used to provide feedback to the language model. The optimization is formulated as:
\begin{equation}\label{eq:RL}
\max_{\pi_{\theta}}  \mathbb{E}_{x\sim \mathcal{D}, y\sim \pi_{\theta}(y \mid x)}\bigl[r_{\phi}(x, y)\bigr] - \beta\mathbb{D}_{\textrm{KL}}\bigl[\pi_{\theta}(y\mid x)\mid \mid \pi_{\mathrm{ref}}(y\mid x)\bigr],
\end{equation}
where $ \pi_{\theta} $ is the policy model parameterized by $ \theta $, 
$ \pi_{\text{ref}} $ is a reference model (e.g., the pre-trained model), 
and $ \beta $ controls the strength of the KL-divergence penalty.

In contrast, DPO~\cite{rafailov2024direct} introduces a simplified framework for preference optimization that bypasses explicit reward modeling. It directly formulates the preference optimization problem as a classification task over preference pairs, eliminating the need for a separate reward model. Compared with RLHF, DPO simplifies the training pipeline while achieving competitive performance in multiple tasks such as dialogue generation. Its objective function is defined as:
\begin{equation}\label{eq:optimum_model}
    \mathcal{L}_\text{DPO}(\pi_{\theta}; \pi_{\mathrm{ref}}) = -\mathbb{E}_{(x, y_w, y_l)\sim \mathcal{D}}\left[\log \sigma \left(\beta \log \frac{\pi_{\theta}(y_w\mid x)}{\pi_{\mathrm{ref}}(y_w\mid x)} - \beta \log \frac{\pi_{\theta}(y_l\mid x)}{\pi_{\mathrm{ref}}(y_l\mid x)}\right)\right],
\end{equation}
where $\beta$ controls the strength of the preference regularization.

\subsection{Motivation: Revisiting DPO}
\subsubsection{Existing Limitations}
\begin{figure}[t]
  \centering
  \vspace{-2mm}
  \begin{subfigure}[b]{0.9\textwidth}
      \centering
      \includegraphics[width=\textwidth]{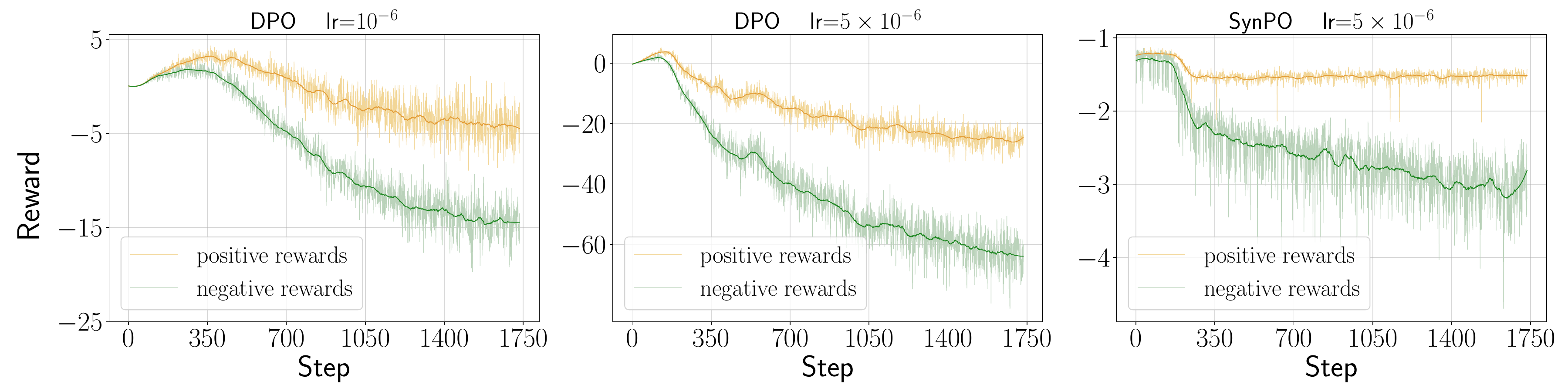}
  \end{subfigure}

  \begin{subfigure}[b]{0.9\textwidth}
      \centering
      \includegraphics[width=\textwidth]{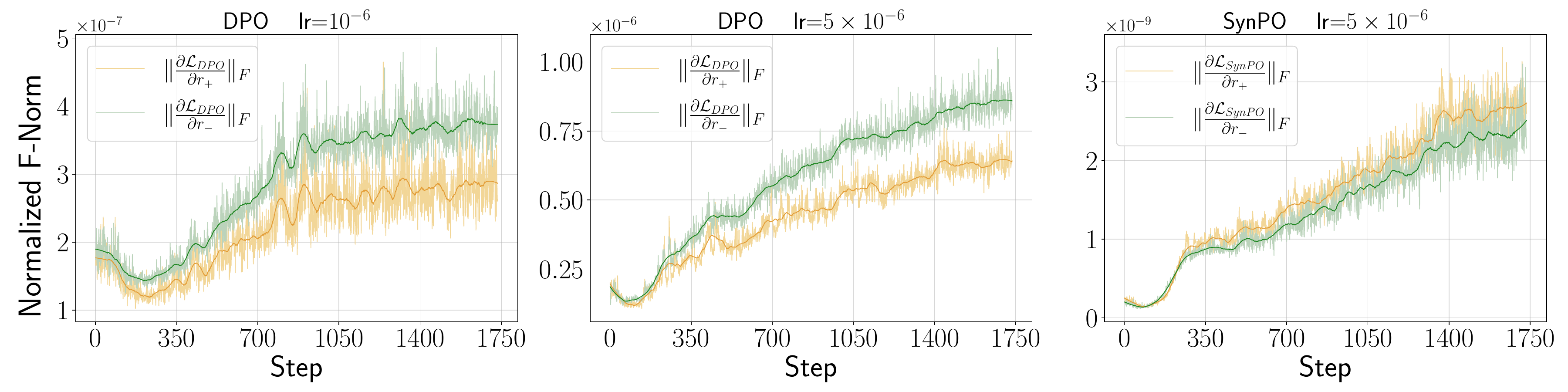}
  \end{subfigure}
  \caption{The evolution of positive and negative rewards and the normalized Frobenius norm of the gradient with respect to positive and negative rewards during training of DPO and SynPO. DPO training undergoes simultaneous decreases in both rewards, with negative preferences dominating the optimization process. Conversely, SynPO mitigates this problem, demonstrating improved performance and stability.}

\label{fig:reward_and_norm}
 \vspace{-5mm}
\end{figure}

Despite  practical success of DPO, several studies have identified limitations within DPO. Specifically, \cite{pal2024smaugfixingfailuremodes} point out that the DPO loss depends solely on the difference between the log-probability ratios of positive and negative preferences. Such a manner suggests that the final loss can decrease even when both are reduced, as long as the negative response decreases more rapidly. \cite{meng2024simposimplepreferenceoptimization} observe preference optimization algorithms tend to
decrease downstream task performance. Moreover, \cite{xiao2025simperminimalistapproachpreference} show that DPO frequently leads to imbalanced updates between positive and negative preferences. As illustrated in Figure~\ref{fig:reward_and_norm}, this imbalance manifests as a concurrent decline in both positive and negative rewards during training, particularly under higher learning rates.

\subsubsection{Theoretical Insights}
From a theoretical perspective, DPO reformulates the original RLHF framework by substituting the reward model with a direct function of the policy model's log-probabilities. By assuming equivalence between the reward models in Eq.~(\ref{eq:reward_model}) and the policy-based formulation in Eq.~(\ref{eq:RL}), DPO reduces the two-step RLHF procedure into a single step. However, this assumption neglects a critical distinction: In  Eq.~(\ref{eq:reward_model}), the reward model constitutes the optimization target with trainable parameters, while in Eq.~(\ref{eq:RL}), the reward model is fixed, and the optimization target is the LLM itself. Obviously, if the reward model in both Eq.~(\ref{eq:reward_model}) and Eq.~(\ref{eq:RL}) were identical and trainable during RLHF training, then in the second phase, due to the KL-divergence penalty relative to the reference model, only the reward model's parameters would typically be updated to minimize the loss, while the LLM's parameters would remain unchanged. This further demonstrates the fundamental difference between the roles of the reward model in Eq.~(\ref{eq:reward_model}) and Eq.~(\ref{eq:RL}), reinforcing that they cannot be treated as equivalent components in optimization; therefore, the aforementioned substitution is theoretically unsound.

We argue that this substitution in the DPO derivation induces a fundamental deviation in the model's optimization objective. In DPO, minimizing the loss is equated with improving the model's ability to rank positive preferences above negative ones, rather than generating higher-quality outputs. Consequently, the model may behave more like a ranking model than a generative one. This deviation from the original goal of RLHF, namely, generating high-reward coherent text, can lead to suboptimal outcomes in terms of language capability. Furthermore, the logarithmic term derived from the KL-divergence constraint in Eq.~(\ref{eq:RL}) fails to serve its intended role in DPO. Due to the derivative properties of the logarithmic function, decreasing reward values require smaller gradient steps than increasing them~\cite{xiao2025simperminimalistapproachpreference}. As a result, the optimizer is incentivized to reduce both positive and negative rewards simultaneously to minimize the overall loss.

\subsubsection{Empirical Observations}
To further investigate this behavior, we analyze the gradient dynamics of the DPO loss with respect to the policy parameters. According to the original DPO paper~\cite{rafailov2024direct}, the gradient is derived as:
\begin{equation*}\label{eq:gradient}
  \resizebox{\textwidth}{!}{$
    \nabla_\theta \mathcal{L}_\text{DPO}(\pi_\theta)=-\beta\mathbb{E}_{(x, y_w, y_l) \sim \mathcal{D}} \bigg[{\sigma(\hat{r}_\theta(x, y_l) - \hat{r}_\theta (x, y_w))}\bigg[{\nabla_\theta\log \pi(y_w \mid x)} - {\nabla_\theta\log\pi(y_l \mid x)}\bigg]\bigg].$}
\end{equation*}

Experimentally, we observe that the normalized Frobenius norm (i.e. $\frac{\Vert A_{m\times n}\Vert_F}{mn}$) of the gradient associated with negative preferences consistently dominates that of positive preferences as training progresses, as shown in Figure~\ref{fig:reward_and_norm}. This indicates that the model updates are primarily driven by the suppression of negative preferences, rather than the promotion of positive ones, a behavior contrary to the intended design of DPO.

However, empirical success has been reported in the original DPO paper and its variants (e.g., DPOP~\cite{pal2024smaugfixingfailuremodes}, IPO~\cite{Azar2023AGT}).These results appear inconsistent with our theoretical findings on the limitations of DPO-style optimization. To understand this discrepancy, we analyze their experimental setups and identified two key factors contributing to the observed performance improvements:

(1) \textbf{Low Learning Rates Mitigate Instability.}
Most DPO-style methods use significantly lower learning rates compared to standard Supervised Fine-Tuning (SFT). As shown in the original DPO paper~\cite{rafailov2024direct}, a learning rate of 1e-6 is used\textemdash much lower than the typical SFT setting of 2e-5 (other variant configurations provided in Tabel \ref{tab:config of synpo variants}). Our gradient-based analysis reveals that under such settings, the magnitude of parameter updates is less than one-tenth of that in standard SFT. This implicitly constrains the model's deviation from its initial state, thereby alleviating the negative effects arising from the deviation of the optimization objective.

(2) \textbf{Preference Discrimination Improves Language Understanding.} 
Encouraging the model to distinguish between positive and negative preferences strengthens its comprehension of human intent and reduces the likelihood of generating hallucinated content. This aligns well with the core motivation behind RLHF, that is, aligning models with preferences. However, such benefits are conditional on maintaining a balanced trade-off between preference discrimination and text generation quality. Specifically, improvements in distinguishing preferences should not come at the cost of deteriorated fluency, coherence, or factual accuracy in generated outputs.

\begin{wrapfigure}{r}{0.35\textwidth}
  \centering
  \vspace{-0.4cm}
  \!\!\!\!\!\!\!\!\includegraphics[width=0.3\textwidth]{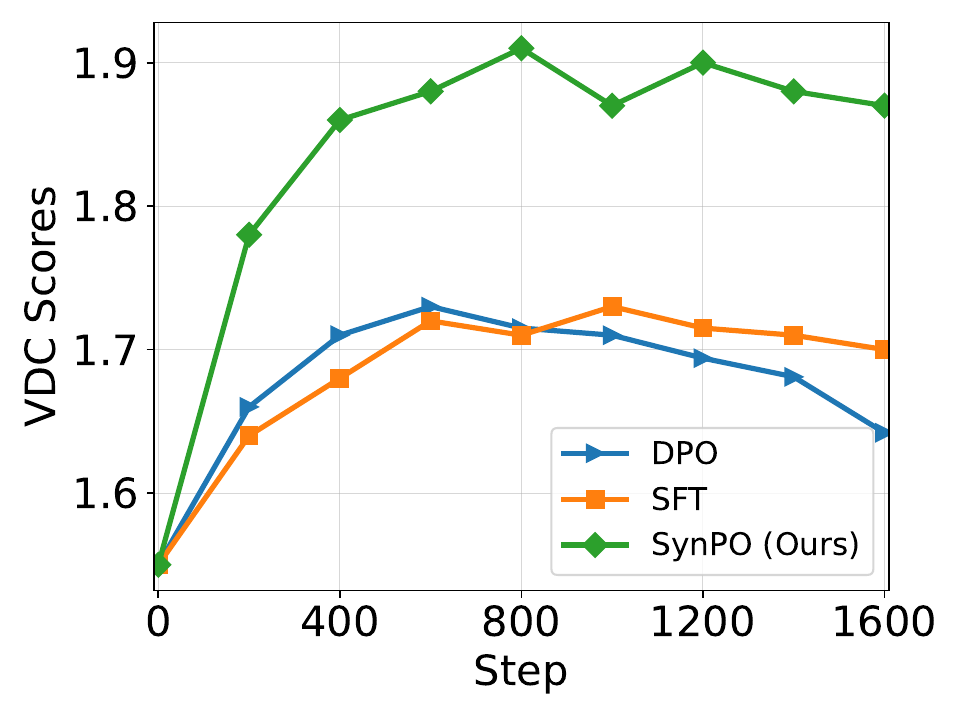}
  \caption{Language capability of different fine-tuning methods.}\label{fig:comparison on linguistic capability}
  \vspace{-0.7cm}
\end{wrapfigure}

Our experiments further support this observation. As shown in Figure~\ref{fig:comparison on linguistic capability}, the DPO-finetuned model exhibits two training phases:
(1) Initial rapid improvement: In the early stages, the model quickly learns to align with preferences and outperforms the SFT baseline, demonstrating the effectiveness of preference-based optimization in enhancing language understanding and generation quality.
(2) Subsequent performance degradation: However, continued training leads to a decline in performance, eventually falling below that of the SFT model. This trend aligns with our theoretical analysis, suggesting that while DPO promotes preference alignment, it may inadvertently weaken the model's language capabilities over time.

\subsection{Solution: Synergizing Descriptiveness and Preference Optimization}
To address the aforementioned limitations, we propose a novel DPO variant named \textbf{SynPO}, which enhances preference alignment while preserving strong language modeling capabilities. Our objective function is formulated as:
\begin{equation}\label{eq:synpo}
  \mathcal{L}_{\mathrm{SynPO}}=-\mathbb{E}_{(x, y_w, y_l)\sim \mathcal{D}}\left[\sigma\left(\alpha \cdot \exp \left(\,\overline{\log S(y_w)}\right)-\alpha \cdot \exp\left(\,\overline{\log S(y_l)\,}\right)\right) + \beta \cdot \overline{S(y_w)} \right],
\end{equation}
where $\alpha$ and $\beta$ are hyperparameters, $y_w$ and $y_l$ denote the positive and negative preferences respectively, $S(y)$ represents a vector of probability values for the entire sequence, where each element corresponds to the probability that the model assigns to each token in the sequence associated with the label $y$, $\log$ is element-wise logarithm for a vector. $(\overline{\,\cdot\,})$ denotes the sample mean, i.e. the average of a vector.
The incorporation of $\mathcal{L}_{\mathrm{SynPO}}$ yields a threefold benefit in model training:

\textbf{(1) Control over Positive and Negative Rewards.}
As mentioned earlier, in standard DPO, the use of $\log$ leads to an improper gradient direction during optimization. Due to the derivative properties of the logarithm, both positive and negative rewards tend to decrease rather than exhibit the desired opposing behavior, that is, one increasing while the other decreases. This tendency causes negative preferences to dominate the optimization process, which is detrimental to the model's ability to learn from preferences. To address this issue, we modify the original DPO reward computation by applying exponential transformations to the positive and negative reward terms. This adjustment effectively alleviates the aforementioned problems and enhances model performance.

\textbf{(2) Empirical Design through Token-level Analysis.}
We conduct an empirical study on token importance using an LLM to score each token based on its semantic contribution to the overall response (full prompts provided in Appendix \ref{sec:full prompts}). As shown in Figure~\ref{fig:token-analysis}, it is observed that tokens with higher semantic importance tend to have lower average log-probabilities. This motivates our use of $\exp \left(\,\overline{\log S(y)}\right)$, which is sensitive to smaller values and better reflects the impact of rare but meaningful tokens on preference learning. Notably, logarithmic averaging amplifies the contribution of smaller values in the vector, while arithmetic averaging is more affected by larger ones. This analysis provides theoretical support for our preference ranking term: the logarithm form amplifies 
meaningful variations in token-level 
confidence, particularly for tokens with low probability, which are often semantically or syntactically critical, while the exponential prevents the simultaneous decrease of positive and negative rewards caused by the logarithm's derivative properties.

\textbf{(3) Explicit Retention of Language Capability.}
In addition to preference ranking, we incorporate
\begin{wrapfigure}{r}{0.4\textwidth}
  \vspace{-0.5cm}
  \!\!\!\!\! \includegraphics[width=0.38\textwidth]{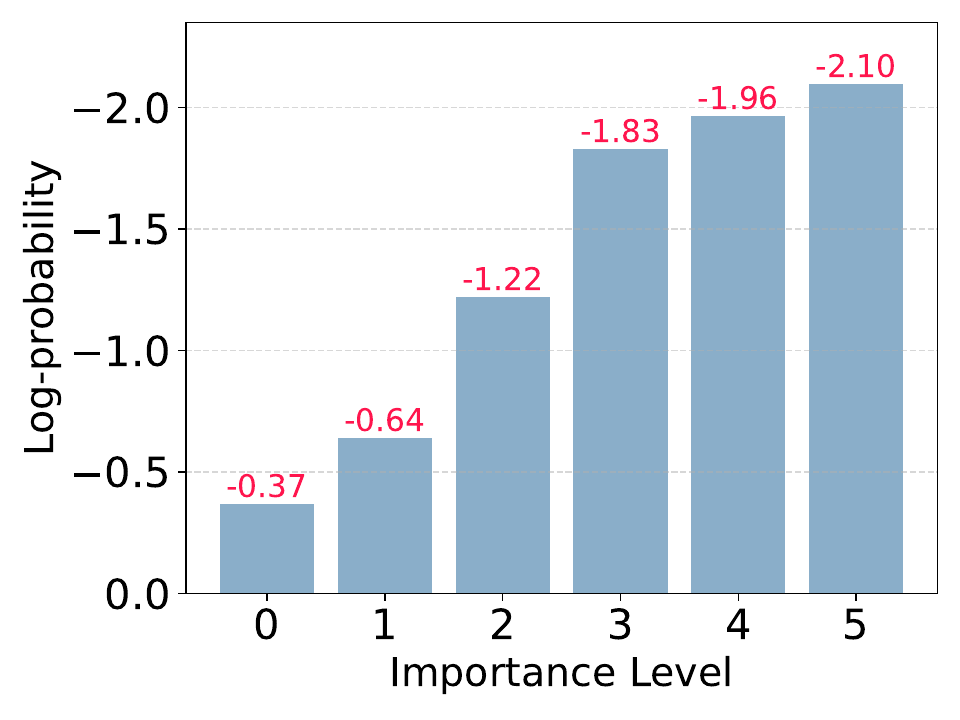}
  \vspace{-0.4cm}
  \caption{Log-probability vs. token importance.}
  \label{fig:token-analysis}
  \vspace{-0.5cm}
\end{wrapfigure}
 an auxiliary term that directly preserves the model's ability to generate fluent and coherent language: $\beta \cdot \overline{S(y_w)}$,
which encourages the model to maintain high token-level fluency across all positions in the preferences. It is worth noting that we avoid the use of log-exponential transformations in the component. Tokens with lower semantic significance (e.g., conjunctions, trailing subwords) often play critical roles in preserving grammatical correctness, and they generally exhibit higher probability values. The scaling nature of logarithm averaging would reduce the impact of these tokens, which are important for fluency and syntactic coherence. As a result, arithmetic averaging is adopted instead.

\textbf{Additional Design Considerations.}
(1) The reference model probability terms used in DPO formula are omitted. Empirically, we find that our optimization process remains stable without them, leading to approximately 20\% faster training compared to standard DPO implementations. (2) The combination of $\alpha$ and $ \sigma(\cdot) $ provides dual benefits: implicit control over preference optimization, and rapid early-stage convergence followed by late-stage stabilization. Due to the derivative properties of sigmoid, increasing $\alpha $ implicitly suppresses excessive preference optimization during training. (3) $\beta$ controls the trade-off between preference ranking and retainment of both semantics and syntax.

\begin{table}[b]
    \centering
    \vspace{-6mm}
    \caption{Ablation study on contrastive decoding (CD) and the self-retrospective strategy (Retro).}
    \label{table:greedy}
    \vspace{1mm}
    \resizebox{0.9\textwidth}{!}{%
    \begin{tabular}{l|c|c|c|c|c|c|c}
        \toprule
        \textbf{Method} & \textbf{Accuracy} & \textbf{Richness} & \textbf{Completeness} &\textbf{Fluency}&\textbf{Dynamics}&\textbf{Coherence}&\textbf{Average}\\
        \midrule
        Baseline & 1.79 & 3.68 & 2.54 & 4.43 & 1.20 & 3.09 & 2.79 \\
        CD only & 1.91 & 3.66 & 2.61 & 4.44 & 1.18 & 3.08 & 2.81\\
        Retro only & 1.78 & 3.86 & 2.55 & 4.54 & 1.27 & 3.19 & 2.87\\
        CD \& Retro & 1.90 & 3.85 & 2.59 & 4.52 & 1.28 & 3.17 & 2.88\\
        \bottomrule
    \end{tabular}}
    
    \vspace{-6mm}
\end{table}

\section{Experiments}\label{sec:experiment}
\subsection{Constructing Long Video Caption Preference Pairs}

\textbf{Enhanced Inference Evaluation.}
The impact of contrastive decoding and the self-retrospective strategy on the quality of generated captions is evaluated. We compare four settings: baseline (no enhancement), contrastive decoding only, self-retrospective only, and the combination of both. Each setting is applied to generate captions on the dataset of VDD~\cite{li2023videochat}, and the outputs are evaluated by an LLM along six dimensions: accuracy, richness, completeness, fluency, dynamics and coherence (full prompts provided in Appendix \ref{sec:full prompts}). As shown in Table~\ref{table:greedy}, combining both strategies achieves the highest scores across all metrics, outperforming either method alone. Specifically:
(1) Contrastive decoding significantly improves accuracy (+6.7\%) and completeness (+2.8\%), indicating its effectiveness in reducing hallucinations and ensuring factual consistency;
(2) The self-retrospective strategy excels in enhancing richness (+5.5\%) and dynamics (+7.6\%), demonstrating its ability to inject more detailed content;
(3) The combined approach retains the strengths of both methods, achieving balanced improvements in all aspects.

\begin{wrapfigure}{r}{0.3\textwidth}
  \centering
  \vspace{-2mm}
  \includegraphics[width=0.3\textwidth]{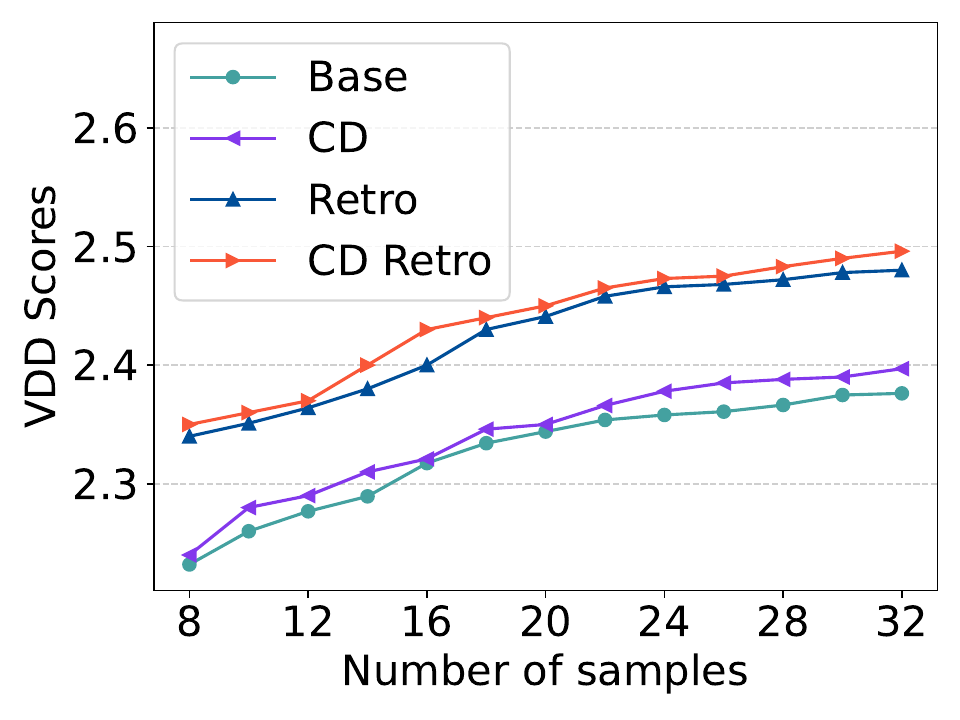}
  \vspace{-5mm}
  \caption{VDD scores of positive preferences constructed via our pipeline under different sampling counts.}
  \label{fig:sampling}
  \vspace{-4mm}
\end{wrapfigure}

\textbf{Ablation Study on Preference Pair Construction.}
Figure~\ref{fig:sampling} illustrates the performance of different augmentation methods across varying sampling counts. Using both techniques simultaneously achieves the best outcomes at identical sampling rates. It is notable that the self-retrospective strategy approximately doubles inference time, while contrastive decoding increases it by 50-75\%. Given this trade-off, we find that employing the self-retrospective strategy with moderately increased sampling yields the most cost-effective approach for preference pair generation.

\begin{table}[t]
    \centering
    \vspace{-2mm}
    \caption{Ablation study on three scoring metrics in the pipeline of constructing preference pairs. 
    }
    \label{tab:criterion}
    \resizebox{1.0\textwidth}{!}{%
    \begin{tabular}{l|c|c|c|c|c|c|c}
        \toprule
         & \textbf{Criterion 1, 2, 3} & \textbf{Criterion 1, 2} & \textbf{Criterion 1, 3} &\textbf{Criterion 2, 3}&\textbf{Criterion 1}&\textbf{Criterion 2}&\textbf{Criterion 3}\\
        \midrule
        \textbf{Positive Preference} & 2.26 & 2.21 & 2.17 & 2.20 & 2.15 & 2.09 & 2.08\\
        \textbf{Negative Preference} & 1.62 & 1.69 & 1.60 & 1.79 & 1.82 & 1.87 & 1.60\\
        \bottomrule
    \end{tabular}}
    \vspace{-1mm}
\end{table}

To assess the contribution of each of our three proposed scoring criteria to the quality of preference pairs, we conduct ablation studies using AuroraCap~\cite{chai2024auroracap} as the base model, generating 10 samples per input with the self-retrospective strategy. Results for different criterion combinations are reported in Table~\ref{tab:criterion}. The results show that all three criteria meaningfully contribute
 to final preference selection. Furthermore, Criterion 1 plays the most critical role in identifying high-quality positive preferences, whereas Criterion 3 demonstrates the strongest effectiveness in distinguishing negative preferences.

\textbf{Downstream Preference Learning.}
To validate the effectiveness of the proposed pipeline, AuroraCap is fine-tuned using several methods, including SFT, DPO~\cite{rafailov2024direct}, and our SynPO. The preference pairs used for fine-tuning were generated from a subset of Sharegpt4video~\cite{chen2024sharegpt4video} through our automated pipeline, with additonal details provided in Appendix \ref{sec:xperimental setup}. Experiments are further conducted across diverse datasets and model configurations.
The resulting models are evaluated on standard video captioning benchmarks (MSR-VTT~\cite{xu2016msr}, VATEX~\cite{wang2020vatexlargescalehighqualitymultilingual}) using CIDEr~(C)~\cite{vedantam2015cider} and METEOR~(M)~\cite{banerjee2005meteor} metrics. In addition, we assess performance on VDC~\cite{chai2024auroracap} and VDD~\cite{li2023videochat} benchmarks, which employ LLM-based evaluation to better measure linguistic richness and factual consistency through their longer, richer reference captions(see Appendix \ref{intro to VDD and VDC} for details). As shown in Table~\ref{tab:workflow-comparison}, all fine-tuned models significantly surpassed their baselines, confirming that our data construction pipeline yields high-quality data.

\begin{table*}[h]
  \centering
  \vspace{-2mm}
  \caption{
  Experimental evaluation on various models (AuroraCap~\cite{chai2024auroracap}, LLaVA1.6-7B-video~\cite{liu2024llavanext}, InterVL2-8B~\cite{wang2025enhancingreasoningabilitymultimodal}), datasets (Sharegpt4video~\cite{chen2024sharegpt4video}, Charades~\cite{sigurdsson2016hollywoodhomescrowdsourcingdata}, Pandas-70M~\cite{chen2024panda70mcaptioning70mvideos}), and fine-tuning approaches (SFT, DPO, SynPO and other variants).
  }
  \label{tab:workflow-comparison}
  \resizebox{\textwidth}{!}{
  \begin{tabular}{@{}lcccccccccc@{}}
      \toprule
      & \multicolumn{5}{c}{\textbf{VDC}} & \multicolumn{1}{c}{\textbf{VDD}} & \multicolumn{2}{c}{\textbf{Vatex}} & \multicolumn{2}{c}{\textbf{MSRVTT}} \\
      \cmidrule(lr){2-6}\cmidrule(lr){7-7}\cmidrule(lr){8-9}\cmidrule(lr){10-11}
      \textbf{AuroraCap fine-tuned with different methods}    & \textbf{Camera}  & \textbf{Short} & \textbf{Background} & \textbf{Main Object} & \textbf{Detail}
      & \textbf{Score}  & \textbf{CIDEr} & \textbf{Meteor} & \textbf{CIDEr} & \textbf{Meteor}\\
      \midrule
      AuroraCap\,(Base) & 1.22  & 1.79 & 1.58  & 1.45
                      & 1.70  &  2.00 & 38.4  & 18.6    
                      & 33.2 & 10.8 \\
      AuroraCap\,(SFT)  & 1.43  & 1.85 & 1.73  & 1.72
                      & 1.91  &  2.18 & 39.2  & 19.0    
                      & 34.0 & 11.2 \\
      AuroraCap\,(DPO~\cite{rafailov2024direct}) 
                      & 1.39  & 1.89 & 1.70  & 1.73
                      & 1.94  &  2.23 & 39.6  & 18.9    
                      & 34.1 & 11.4 \\
      AuroraCap\,(DPOP~\cite{pal2024smaugfixingfailuremodes}) 
                      & 1.55  & 1.88 & 1.78  & 1.79
                      & 1.96  &  2.30 & 41.1  & 19.3    
                      & 34.8 & \textbf{11.5} \\
      AuroraCap\,(IPO~\cite{Azar2023AGT}) 
                    & 1.44  & 1.83 & 1.74  & 1.76
                      & 1.95  &  2.21 & 39.8  & 19.1   
                      & 34.3 & 11.3 \\
      AuroraCap\,(KTO~\cite{Ethayarajh2024KTOMA}) 
                    & 1.53  & 1.88 & 1.73  & 1.78
                      & 1.97  &  2.27 & 40.1  & 19.0    
                      & 34.3 & 11.4 \\
      AuroraCap\,(CPO~\cite{xu2024contrastive}) 
                    & 1.42  & 1.81 & 1.75  & 1.72
                      & 1.94  &  2.25 & 40.8  & 18.9    
                      & 34.1 & 11.2 \\
      AuroraCap\,(SimPO~\cite{meng2024simposimplepreferenceoptimization})               & 1.52  & 1.83 & 1.76  & 1.75
                      & 1.96  &  2.26 & 40.2  & 19.1    
                      & 34.5 & 11.3 \\
      AuroraCap\,(SynPO-v1) & 1.72  & 1.89 & 1.87  & 1.83
                      & 2.02  &  2.35 & 42.1  & 19.5    
                      & 35.2 & \textbf{11.5} \\
      AuroraCap\,(SynPO-v2) & 1.74  & 1.93 & 1.88  & \textbf{1.87}
                      & 2.03  &  2.37 & 42.3  & 19.5    
                      & \textbf{35.4} & 11.4 \\
      AuroraCap\,(SynPO-v3) & 1.75  & 1.91 & 1.90  & 1.84
                      & \textbf{2.05}  &  2.38 & 42.3  & \textbf{19.6}    
                      & 35.3 & 11.3 \\
      AuroraCap\,(SynPO-v4) & 1.48  & 1.87 & 1.80  & 1.75
                      & 1.98  &  2.25 & 40.9  & 19.2    
                      & 34.6 & 11.2 \\
     AuroraCap\,(SynPO-v5) & 1.57  & 1.78 & 1.82  & 1.77
                      & 1.97  &  2.29 & 41.2  & 19.3    
                      & 34.5 & 11.2 \\
     \rowcolor{gray!20}\textbf{AuroraCap\,(SynPO)} & \textbf{1.78}  & \textbf{1.94} &                         \textbf{1.91}  & \textbf{1.87}
                      & 2.04  &  \textbf{2.43} & \textbf{42.5}  
                      &\textbf{19.6} & \textbf{35.4} & \textbf{11.5} \\
      \midrule
      \textbf{AuroraCap fine-tuned on more dataset} &&&&&&&&&&\\
      Charades\,(SFT) & 1.40  & 1.87 & 1.68  & 1.75
                      & 1.90  &  2.21 & 39.5  & 19.2    
                      & 33.8 & 11.3 \\
      Charades\,(DPO) & 1.41  & 1.88 & 1.66  & 1.73
                      & 1.92  &  2.19 & 39.3  &  19.2  
                      & 34.0 & 11.2 \\
      \rowcolor{gray!20} \textbf{Charades\,(SynPO}) & 1.75  & \textbf{1.95} & 1.88  & 1.88
                      & 2.02  &  2.41 & 42.2  & 19.6    
                      & \textbf{35.2} & 11.5 \\
      Pandas-70M\,(SFT) & 1.45  & 1.86 & 1.71  & 1.74
                      & 1.92  &  2.22 & 39.8  & 19.2    
                      & 34.1 & 11.2 \\
      Pandas-70M\,(DPO) & 1.44  & 1.88 & 1.75  & 1.75
                      & 1.93  &  2.25 & 40.5  & 19.3    
                      & 33.9 & 11.1 \\
      \rowcolor{gray!20}\textbf{Pandas-70M\,(SynPO)} & \textbf{1.79}  & 1.94 & \textbf{1.91}  &                      \textbf{1.89} & \textbf{2.06} & \textbf{2.44} &                     \textbf{42.6}  & \textbf{19.7}   
                      & 35.0 & \textbf{11.6} \\

      \midrule
      \textbf{Models fine-tuned on Sharegpt4Video} &&&&&&&&&&\\
      LLaVA1.6-7B-video     & 1.14  & 1.75 & 1.63  & 1.41
                      & 1.63  &  1.89 & 33.5  & 16.8    
                      & 29.6 & 10.1 \\
      LLaVA1.6-7B-video\,(SFT) & 1.37  & 1.83 & 1.72  & 1.66
                      & 1.88  &  2.13 & 34.7  & 17.3    
                      & 30.8 & 10.5 \\
      LLaVA1.6-7B-video\,(DPO) & 1.45  & 1.87 & 1.71  & 1.72
                      & 1.90  &  2.19 & 35.3  & 17.3    
                      & 30.7 & 10.5 \\
      \rowcolor{gray!20} \textbf{LLaVA1.6-7B-video\,(SynPO)} & 1.74  & 1.90 & 1.94  & 1.85
                      & 2.00  &  2.36 & 37.3  & 18.1    
                      & 32.1 & 10.9 \\
      InternVL2-8B&     1.26  & 1.83 & 1.66  & 1.64
                      & 1.69  &  2.15 & 39.2  & 19.2    
                      & 33.9 & 10.9 \\
      InternVL2-8B\,(SFT) & 1.46  & 1.88 & 1.77  & 1.83
                      & 1.93  &  2.26 & 40.3  & 19.7    
                      & 34.7 & 11.5 \\
      InternVL2-8B\,(DPO) & 1.44  & 1.92 & 1.82  & 1.89
                      & 1.91  &  2.31 & 40.7  & 19.6    
                      & 35.2 & 11.3 \\
      \rowcolor{gray!20} \textbf{InternVL2-8B\,(SynPO)} & \textbf{1.80}  & \textbf{1.96} &                            \textbf{1.95}  & \textbf{1.97}
                      & \textbf{2.04} & \textbf{2.48} & \textbf{42.8}  & \textbf{20.1} & \textbf{36.3} & \textbf{11.7} \\
      \bottomrule
  \end{tabular}}
  \vspace{-3mm}
\end{table*}

\subsection{Experiments of SynPO}
\subsubsection{Comparison with DPO and Various Variants}
As shown in Table~\ref{tab:workflow-comparison}, SynPO and several of its variants, as well as DPO variants, are compared under identical training settings and evaluation metrics (as defined in Section 5.1 Downstream 
\setlength{\tabcolsep}{4pt}
\begin{wraptable}{r}{0.5\textwidth}
  \vspace{-1mm}
  \centering
    \caption{SynPO variants.}
  \label{tab:SynPO-variants}
  \vspace{-2mm}
  \resizebox{0.5\textwidth}{!}{%
  \begin{tabular}{lc}
  \toprule 
  \textbf{Method} & \textbf{Objective Function} \\ \midrule
  SynPO-v1 & $-\sigma\left(\alpha \cdot \exp \left(\overline{\log S(y_w)}\right)-\alpha \cdot \exp \left(\overline{\log S(y_l)}\right)\right)$ \\ \midrule
  SynPO-v2 & $-\sigma\left(\alpha \cdot \exp \left(\overline{\log S(y_w)}\right)-\alpha \cdot \exp \left(\overline{\log S(y_l)}\right)\right)-\beta \, \overline{\log S(y_w)}$ \\ \midrule 
  SynPO-v3 & $-\sigma\left(\alpha \cdot \overline{S(y_w)}-\alpha \cdot\overline{S(y_l)}\right)-\beta \, \overline{S(y_w)}$ \\ \midrule 
  SynPO-v4 & $-\sigma\left(\alpha \cdot \overline{S(y_w)}-\alpha \cdot \overline{S(y_l)}\right)-\beta \, \overline{\log S(y_w)}$ \\  \midrule 
  SynPO-v5 &  $-\left(\alpha \cdot \exp \left(\overline{\log S(y_w)}\right)-\alpha \cdot \exp \left(\overline{\log S(y_l)}\right)\right)-\beta \, \overline{S(y_w)}$ \\ 
  \bottomrule
    \vspace{-2mm}
  \end{tabular}}

  \vspace{-0.7cm}
\end{wraptable}
Preference Learning). The results indicate that SynPO typically outperforms other variants, including those detailed in Table~\ref{tab:SynPO-variants}. Comparisons with various SynPO variants confirm that our modifications to the formula, specifically the incorporation of $\sigma(\cdot)$, logarithmic and exponential functions, yield significant performance improvements, validating the effectiveness and optimality of our approach.

\subsubsection{Effectiveness in NLP Domain}

\textbf{Training Recipe.} Training experiments are conducted using Llama3-8B~\cite{llama3modelcard} (Base and Instruct) and Mistral-7B~\cite{Jiang2023Mistral7} (Base and Instruct). For both Llama3-8B-Base and Mistral-7B-Base, we employ a training pipeline  \cite{tunstall2023zephyr}. First, we train a base model on the UltraChat-200k dataset~\cite{Ding2023EnhancingCL} to obtain an SFT model. Then, we perform preference optimization on the UltraFeedback dataset~\cite{Cui2024UltraFeedbackBL} using the SFT model as the starting point. For Llama3-8B-Instruct and Mistral-7B-Instruct, we implement an on-policy evaluation strategy following SimPO~\cite{meng2024simposimplepreferenceoptimization}. Specifically, prompts from UltraFeedback are used to regenerate positive and negative preference pairs via SFT models.  For each prompt, five responses are sampled from the SFT model and rank them using PairRM (LLM-Blender)~\cite{jiang2023llm}. The highest-ranked response is selected as the positive preference, and the lowest-ranked response is designated as the negative one.

\textbf{Evaluation Benchmark.} Building upon recent methodologies in preference-based fine-tuning~\cite{rafailov2024direct,tunstall2023zephyr}, we evaluate model performance using standardized frameworks. These include both versions of the HuggingFace Open LLM Leaderboard~\cite{eval-harness}, summarized in Table~\ref{tab:nlp_tasks}, and comprehensive instruction-following benchmarks (AlpacaEval2 and MT-Bench) as reported in Table~\ref{tab:MT-bench}. 
Detailed descriptions of evaluation tasks and procedures are provided in Appendix \ref{sec: evaluation_benchmark}.

\setlength{\tabcolsep}{2pt}
\begin{table*}[t]
    \centering
    \small 
    \caption{AlpacaEval2~\cite{AlpacaEval} and MT-Bench~\cite{zheng2023judging} results under the four settings. LC and WR denote length-controlled and raw win rate, respectively. }
    \label{tab:MT-bench}
    \vspace{-1mm}
    \resizebox{\textwidth}{!}{
    \begin{tabular}{l ccc ccc ccc ccc}
    \toprule
    \multirow{3}{*}{\textbf{Method}} 
    & \multicolumn{3}{c}{\textbf{Mistral-7B-Base}} 
    & \multicolumn{3}{c}{\textbf{Mistral-7B-Instruct}} 
    & \multicolumn{3}{c}{\textbf{Llama3-8B-Base}} 
    & \multicolumn{3}{c}{\textbf{Llama3-8B-Instruct}} 
    \\ 
    \cmidrule(lr){2-4}\cmidrule(lr){5-7}\cmidrule(lr){8-10}\cmidrule(lr){11-13}
    & \multicolumn{2}{c}{\textbf{AlpacaEval 2}} 
    & \multicolumn{1}{c}{\textbf{MT-Bench}} 
    & \multicolumn{2}{c}{\textbf{AlpacaEval 2}} 
    & \multicolumn{1}{c}{\textbf{MT-Bench}} 
    & \multicolumn{2}{c}{\textbf{AlpacaEval 2}} 
    & \multicolumn{1}{c}{\textbf{MT-Bench}} 
    & \multicolumn{2}{c}{\textbf{AlpacaEval 2}} 
    & \multicolumn{1}{c}{\textbf{MT-Bench}} \\
    \cmidrule(lr){2-3}\cmidrule(lr){4-4} \cmidrule(lr){5-6} \cmidrule(lr){7-7} \cmidrule(lr){8-9}\cmidrule(lr){10-10}\cmidrule(lr){11-12}\cmidrule(lr){13-13}
    & {\scriptsize \bf LC (\%)} 
    & {\scriptsize \bf WR (\%)} 
    & {\scriptsize \bf GPT-4} 
    & {\scriptsize \bf LC (\%)}  
    & {\scriptsize \bf WR (\%)} 
    & {\scriptsize \bf GPT-4} 
    & {\scriptsize \bf LC (\%)}  
    & {\scriptsize \bf WR (\%)} 
    & {\scriptsize \bf GPT-4} 
    & {\scriptsize \bf LC (\%)}  
    & {\scriptsize \bf WR (\%)} 
    & {\scriptsize \bf GPT-4} \\
    \midrule
    SFT 
    &  8.4 & 6.2 & 6.3 & 17.1 & 14.7 & 7.5 
    &  6.2 & 4.6 & 6.6 & 26.0 & 25.3 & 8.1
    \\
    \midrule
    DPO~\cite{rafailov2024direct}  
    & 15.1 & 12.5 & 7.3 & 26.8 & 24.9 & 7.6 
    & 18.2 & 15.5 & 7.7 & 40.3 & 37.9 & 8.0
    \\
    DPOP~\cite{pal2024smaugfixingfailuremodes}
    & 16.1 &  12.8 & 7.4 & 27.1 & 24.6 & 7.7
    & 16.7 & 14.3 & 7.6 & 45.2 & 39.1 & 8.2
    \\
    IPO~\cite{Azar2023AGT}  
    & 11.8 & 9.4 & 7.2 & 20.3 & 20.3 & 7.8 
    & 14.4 & 14.2 & 7.4 & 35.6 & 35.6 & \textbf{8.3}
    \\
    KTO~\cite{Ethayarajh2024KTOMA}  
    & 13.1 & 9.1 & 7.0 & 24.5 & 23.6 & 7.7 
    & 14.2 & 12.4 & \textbf{7.8} & 33.1 & 31.8 & 8.2
    \\
    CPO~\cite{xu2024contrastive} 
    & 9.8 &  8.9 &  6.8 & 23.8 & 28.8 & 7.5 
    & 10.8 & 8.1 & 7.4 & 28.9 & 32.2 & 8.0
    \\
    SimPO~\cite{meng2024simposimplepreferenceoptimization} 
    & 21.5 & 20.8 & 7.3 & 32.1 & 34.8 & 7.6 
    & 22.0 & 20.3 & 7.7 & 44.7 & 40.5 & 8.0
    \\
    \rowcolor{gray!20}\textbf{SynPO} & \textbf{22.9} & \textbf{22.1} &  \textbf{7.7} & \textbf{37.9} & \textbf{39.8} &  \textbf{7.9} & \textbf{25.7} & \textbf{23.1} & 7.7 & \textbf{49.0} & \textbf{46.2} & \textbf{8.3}\\
    \bottomrule
    \end{tabular}}
\end{table*}

\textbf{Analysis.} Comparative evaluations conducted across multiple models and diverse tasks clearly show the superiority of our proposed method over alternative optimization methods. SynPO consistently delivers favorable results on both preference tasks and downstream applications, suggesting that it effectively enhances the model's capacity to discern between positive and negative preferences, while simultaneously improving its general language understanding and generation abilities.

\begin{table*}[t]
    \centering
    \small
    \caption{Evaluation results on various tasks from the Huggingface Open Leaderboards~\cite{open-llm-leaderboard-v1,open-llm-leaderboard-v2} show that our SynPO achieves superior or comparable performance to other.}
    \label{tab:nlp_tasks}
    \resizebox{1\textwidth}{!}{%
    \begin{tabular}{cl*{9}{c}}
    \toprule[1pt]
     & \textbf{Method} &  \textbf{MMLU-PRO} & \textbf{IFEval} & \textbf{BBH} & \textbf{HellaSwag} & \textbf{WinoGrande} & \textbf{TruthfulQA} & \textbf{GSM8K} & \textbf{ARC-C} & \textbf{Average}\\
    \midrule[0.5pt]
    \multirow{7}{*}{\parbox[t]{1.7cm}{\centering \textbf{Mistral-7B} \\ \textbf{Base} }} 
    & DPO  & 35.68 & 29.18 & 26.85 & 81.40 & 76.66 & 48.72 & 53.47 & 54.32  & 50.79\\
    & DPOP  & 36.12 & 30.75 & 26.24 & 80.49 & 76.23 & 49.07 & 54.32 & 54.73 &  50.99 \\
    & IPO  & 34.87 & 25.52 & 25.59 & 79.15 & 74.15 & 47.25 & 54.14  & 53.84  & 49.31 \\
    & KTO & 35.51 & 27.03 & 27.66 & \textbf{81.75} & \textbf{77.17} & 48.34 & 54.51  & 54.37 & 50.79 \\
    & CPO & 34.04 & 26.32 & 27.05& 80.45 & 75.15 & 49.15 & 53.06 & 54.53 & 49.97 \\
    & SimPO &  35.13 & 29.63 & 26.94 & 81.03 & 76.68 & 49.49 & 53.21 &  53.63 &  50.72 \\
    & \textbf{SynPO} & \textbf{37.84} & \textbf{30.83} & \textbf{28.99} & 81.26 & 77.14 & \textbf{50.58} & \textbf{54.95} & \textbf{55.24} & \textbf{52.10} \\
     \midrule[0.8pt]
    \multirow{7}{*}{\parbox[t]{1.7cm}{\centering \textbf{LLama3-8B} \\ \textbf{Base} }}  
    & DPO  & 36.53 & 30.97 & 27.34 & 80.02 & \textbf{75.46} & 49.15 & 53.45 & 54.61 & 50.94   \\
    & DPOP  & 36.92 & 29.38 & 26.96 & 79.62 & 74.53 & 50.12 & 52.35 & 53.20 &  50.38 \\
    & IPO & 35.47 & 25.76 & 27.01 & 79.48 & 74.87 & 48.03 & 51.76 & 51.83 &  49.28 \\
    & KTO & 35.89 & 28.88 & 27.09 & 78.53 & 75.12 & 51.67 & 52.45 & 52.84 &  50.31 \\
    & CPO & 36.18 & 26.86 & 28.14 & 79.42 & 73.32 & 49.04 & 51.52 & 54.11 &  49.82 \\
    & SimPO & 36.13 & 27.73 & 26.88 & 78.13 & 73.46 & 49.15 & 53.45 & 52.61 &   49.69   \\
    & \textbf{SynPO} & \textbf{38.03} & \textbf{31.42} & \textbf{29.03} & \textbf{80.88} & 74.55 & \textbf{53.04} & \textbf{54.81} & \textbf{55.54} & \textbf{52.16}\\
         \midrule[0.8pt]
    \multirow{7}{*}{\parbox[t]{1.7cm}{\centering \textbf{Mistral-7B} \\ \textbf{Instruct}}}  
    & DPO  & 39.53 & 33.72 & 29.34 & 83.22 & 78.46 & 51.15 & 54.22 & 60.04 &   53.71   \\
    & DPOP  & 39.69 & 34.53 & 29.24 & 82.04 & \textbf{80.13} & 52.95 & 53.65 & 59.90 & 54.02 \\
    & IPO & 38.75 & 31.85 & 30.21 & 81.61 & 79.55 & 52.02 & 52.42 & 58.31 & 53.09 \\
    & KTO & \textbf{40.46} & 34.02 & 30.62 & 80.34 & 78.19 & 52.77 & 53.35 & 59.80 & 53.69 \\
    & CPO & 38.85 & 27.81 & 32.66 & 80.01 & 79.15 & 50.28 & 52.28 &  58.74 & 52.47 \\
    & SimPO & 39.10 & 29.52 & 32.70 & 82.04 & 78.71 & 52.19 & 54.25 & 59.69 & 53.52 \\
    & \textbf{SynPO} & 40.08 & \textbf{35.84} & \textbf{32.87} & \textbf{83.76} & 79.92 & \textbf{54.51} & \textbf{55.11} & \textbf{60.43} & \textbf{55.32}\\
      \midrule[0.8pt]
    \multirow{7}{*}{\parbox[t]{1.7cm}{\centering \textbf{LLama3-8B} \\ \textbf{Instruct}}} 
    & DPO  & 41.32 & 34.54 & 31.29 & 82.85 & 78.22 & 52.81 & 54.83 & 59.76 &   54.45   \\
    & DPOP  & 41.89 & \textbf{36.51} & 30.55 & 82.52 & 79.10 & 52.29 & 53.57 & 59.26 & 54.46 \\
    & IPO  & 40.97 & 33.27 & 30.31 & 81.95 & 78.58 & 51.02 & 54.23 & 59.95 & 53.78 \\
    & KTO & 41.70 & 34.12 & 31.15 & 82.70 & 77.10 & 53.63 & 54.01 & 60.57 & 54.37 \\
    & CPO & 39.56 & 35.08 & 30.51 & 81.08 & 76.81& 52.75 & 53.40 & 58.29 & 53.44 \\
    & SimPO & 40.09 & 35.05 & 30.95 & 82.29 & 77.15 & 53.16  & 54.72 & \textbf{61.24} & 54.33 \\
      & \textbf{SynPO} & \textbf{42.08} & 36.06 & \textbf{31.98} & \textbf{83.19} & \textbf{79.71} & \textbf{54.35} & \textbf{55.37} & 60.61 & \textbf{55.42}\\
    \bottomrule[1pt]
    \end{tabular}}
\end{table*}

\section{Conclusion}
\label{sec:conclusion}

This research addresses two fundamental challenges hindering fine-grained video captioning: the lack of scalable, high-quality preference data and the practical limitations of standard DPO. To generate preference data efficiently, we develop an automated pipeline requiring neither human annotation nor access to stronger VLMs. Concurrently, our theoretical and empirical analysis reveals DPO's core issues: excessive focus on negative examples and deviation from ranking optimization. Our solution, SynPO, counteracts these by rebalancing preference signals and incorporating generation-preserving terms, leading to improved language capability and training efficiency. SynPO's effectiveness and broad applicability are validated through extensive experiments on diverse video captioning and NLP benchmarks, where it consistently outperforms existing methods.

\clearpage
\printbibliography

\clearpage
\newpage
\clearpage
\appendix

\section{The Evaluation Benchmarks}
\label{sec: evaluation_benchmark}
\textbf{MMLU-PRO.}~\cite{wang2024mmluprorobustchallengingmultitask} It is a robust benchmark for evaluating cross-disciplinary reasoning in LLMs, comprising 12,000 complex questions spanning STEM, humanities, and professional domains, with ten answer options per question to minimize random guessing and emphasize analytical depth. It integrates problems from diverse sources (e.g., original MMLU, TheoremQA, SciBench), employs chain-of-thought reasoning requirements, and demonstrates enhanced robustness to prompt variations, as evidenced by leading models achieving ~71\% accuracy while highlighting significant performance gaps compared to earlier benchmarks.

\textbf{IFEval.}~\cite{zhou2023instructionfollowingevaluationlargelanguage} It is a benchmark dataset designed to evaluate the in-context learning and few-shot reasoning capabilities of LLMs across diverse NLP tasks, featuring carefully curated prompts and annotations to assess performance under varying input conditions and task complexities.

\textbf{BBH.}~\cite{suzgun2022challengingbigbenchtaskschainofthought} Big Bench Hard is a benchmark dataset designed to evaluate the cross-domain reasoning capabilities of LLMs, comprising 23 high-difficulty tasks that emphasize multi-step logical deduction, attention control, and memory retention, with a focus on few-shot learning scenarios and the application of chain-of-thought (CoT) reasoning to challenge models beyond their standard performance thresholds.

\textbf{HellaSwag.}~\cite{zellers2019hellaswagmachinereallyfinish} It is a NLP benchmark designed to evaluate machine commonsense reasoning and contextual understanding, featuring more than 100,000 context-rich question-answer pairs generated via crowdsourcing and adversarial filtering, with a focus on challenging models to infer plausible continuations of text beyond superficial pattern matching.

\textbf{WinoGrande.}~\cite{sakaguchi2019winograndeadversarialwinogradschema} It is a benchmark dataset designed to evaluate the commonsense reasoning and pronoun disambiguation capabilities of LLMs, extending the Winograd Schema Challenge by introducing 44,000 context-dependent questions with multiple-choice answers that require resolving ambiguous references through deep contextual understanding and implicit world knowledge.

\textbf{TruthfulQA.}~\cite{lin-etal-2022-truthfulqa} It is designed to evaluate the factual accuracy and truthfulness of LLMs, comprising 817 adversarially crafted zero-shot questions across 38 topics with verified true/false answers, emphasizing the model's ability to avoid generating false statements through rigorous human-validated sources and challenging high-probability training-distribution biases.

\textbf{GSM8K.}~\cite{cobbe2021trainingverifierssolvemath} It is a benchmark dataset designed to evaluate the multistep arithmetic reasoning capabilities of NLP models, comprising 8,500 high-quality grade-school-level math word problems that require 2–8 sequential operations using basic arithmetic, with answers presented in annotated natural language formats to facilitate both model training and rigorous assessment of mathematical problem-solving robustness.

\textbf{ARC-C.}~\cite{clark2018thinksolvedquestionanswering} The ARC dataset consists of 7,787 science questions, all non-diagram, multiple choice (tpically 4-way multiple choice). They are drawn from a variety of sources, and sorted into a challenge set of 2,590 ``hard" questions (those that both a retrieval and a co-occurrence method fail to answer correctly) and an easy set of 5,197 questions. Questions vary in their target student grade level (as assigned by the examiners who authored the questions), ranging from 3rd grade to 9th.

\section{Details of DPO Variants}

Furthermore, we provides a detailed introduction below to state-of-the-art baselines for preference fine-tuning, with an emphasis on the usage of hyperparameters in their objective functions which are listed in Table~\ref{tab:DPO variants}.

\begin{table*}[!h]
  \vspace{-2mm}
  \caption{Various preference optimization objectives and search spaces for hyperparameters.}
  \label{tab:DPO variants}
  \centering
  \small
  \resizebox{\textwidth}{!}{%
  \begin{tabular}{llll}
  \toprule 
  \textbf{Method} & \textbf{Objective} & \textbf{Hyperparameter} \\ \midrule  
  DPO & $-\log \sigma \left( \beta \log \frac{\pi_\theta(y_w|x)}{\pi_{\text{ref}}(y_w|x)} - \beta \log \frac{\pi_\theta(y_l|x)}{\pi_{\text{ref}}(y_l|x)}\right)$ & $\beta \in \{0.01, 0.05, 0.1\}$ \\ \midrule 
  IPO & $ \left( \log \frac{\pi_\theta(y_w|x)}{\pi_{\text{ref}}(y_w|x)} - \log \frac{\pi_\theta(y_l|x)}{\pi_{\text{ref}}(y_l|x)} - \frac{1}{2\tau} \right)^2$ & $\tau \in \{0.01, 0.1, 0.5, 1.0\}$ \\  \midrule 
  CPO &  $-\log \sigma  \left(\beta \log \pi_\theta(y_w|x) - \beta \log \pi_\theta(y_l|x) \right) - \lambda \log \pi_\theta (y_w|x)$ & $\lambda = 1.0, \,\, \beta \in \{0.01, 0.05, 0.1\}$ &\\ \midrule
  DPOP& $-\left[ \log \sigma \left( \beta \left( \log \frac{\pi_{\theta}(y_w | x)}{\pi_{\text{ref}}(y_w | x)} - \log \frac{\pi_{\theta}(y_l | x)}{\pi_{\text{ref}}(y_l | x)} 
 - \lambda \cdot \max \left( 0, \log \frac{\pi_{\text{ref}}(y_w | x)}{\pi_{\theta}(y_w | x)} \right) \right) \right) \right]$  &
 $\beta \in \{0.5,0.1,0.2,0.3\}$, $\lambda \in \{5,10,25,50\}$
 & \\ \midrule
  \multirow{2}{*}{KTO} & $\begin{aligned}[t] &-\lambda_w \sigma \left( \beta \log \frac{\pi_\theta(y_w|x)}{\pi_{\text{ref}}(y_w|x)} - z_{\text{ref}} \right) +  \lambda_l \sigma \left( z_{\text{ref}} - \beta \log \frac{\pi_\theta(y_l|x)}{\pi_{\text{ref}}(y_l|x)} \right), \\ &\text{where } z_{\text{ref}} = \mathbb{E}_{(x, y) \sim \mathcal{D}} \left[\beta \text{KL}\left( \pi_\theta(y|x) \| \pi_{\text{ref}}(y|x) \right) \right] \end{aligned}$ & $\lambda_l = \lambda_w = 1.0$, $\beta \in \{0.01, 0.05, 0.1\}$ &\\ \midrule 
  \multirow{2}{*}{SimPO} & \multirow{2}{*}{$-\log \sigma  \left( \frac{\beta}{|y_w|} \log \pi_\theta(y_w|x) - \frac{\beta}{|y_l|} \log \pi_\theta(y_l|x) - \gamma \right)$} & $\beta \in \{2.0, 2.5\}$ \\
  & & $\gamma \in \{0.3, 0.5, 1.0, 1.2, 1.4, 1.6\}$ &\\ \midrule
  SynPO & $-\left[\sigma\left(\alpha \cdot \exp \left(\,\overline{\log S(y_w)}\right)-\alpha \cdot \exp\left(\,\overline{\log S(y_l)\,}\right)\right) + \beta \cdot \overline{S(y_w)} \right]$ &
  $\alpha \in \{20,30,50\}$, $\beta \in \{0.1,0.2,0.3\}$
  & \\ 
  \bottomrule
  \end{tabular}}
\end{table*}

\textbf{DPO.} Direct Preference Optimization~\cite{rafailov2024direct} uses log-likelihood differences to implicitly represent the reward function, eliminating the need for explicit reward model like RLHF. 
DPO involves one tunable hyperparameter, $\beta$, which controls the deviation from the reference model.

\textbf{IPO.} Identity Preference Optimization~\cite{azar2024general} minimizes a squared loss regression problem by defining an alternative reward function, avoiding unstable RL training. 
IPO involves one hyperparameter, $\beta$, to adjust the reward margin.

\textbf{CPO.} Contrastive Preference Optimization~\cite{xucontrastive} uses log-likelihood as the reward and is trained alongside a Supervised Fine-Tuning (SFT) objective. 
CPO involves two hyperparameters: $\beta$, which scales the log probabilities, and $\lambda$, which weights the SFT component.

\textbf{SimPO.} Simple Preference Optimization~\cite{meng2024simposimplepreferenceoptimization} eliminates the need for a reference model and optimizes a length-regularized probability of response pairs. 
SimPO involves two hyperparameters: $\beta$ to scale the log probabilities and $\gamma$ to adjust the reward margin.

\textbf{KTO.} Kahneman-Tversky Optimization~\cite{ethayarajh2024kto} learns from non-paired preference data. 
KTO involves three hyperparameters: $\beta$, which controls the deviation from the reference model; $\lambda_w$ and $\lambda_l$, which weight the preference components for winning and losing responses, respectively.

\textbf{DPOP.} DPO-Positive~\cite{pal2024smaugfixingfailuremodes} adds a new term to the loss which leads every token to be incentivised toward the preferred completion.

\begin{table*}[!h]
  \vspace{-1em}
  \caption{The SynPO variants and search spaces for hyperparameters.}
  \label{tab:config of synpo variants}
  \centering
  \small
  \resizebox{\textwidth}{!}{%
  \begin{tabular}{llll}
  \toprule 
  \textbf{Method} & \textbf{Objective} & \textbf{Hyperparameter} \\ \midrule  
  SynPO-v1 & $-\sigma\left(\alpha \cdot \exp \left(\overline{\log S(y_w)}\right)-\alpha \cdot \exp \left(\overline{\log S(y_l)}\right)\right)$ 
  & $\alpha \in \{20,30,50\}$
  \\ \midrule
  SynPO-v2 & $-\sigma\left(\alpha \cdot \exp \left(\overline{\log S(y_w)}\right)-\alpha \cdot \exp \left(\overline{\log S(y_l)}\right)\right)-\beta \, \overline{\log S(y_w)}$ 
  & $\alpha \in \{20,30,50\}$, $\beta \in \{0.05,0.1,0.2,0.3\}$
  \\ \midrule 
  SynPO-v3 & $-\sigma\left(\alpha \cdot \overline{S(y_w)}-\alpha \cdot\overline{S(y_l)}\right)-\beta \, \overline{S(y_w)}$ 
  & $\alpha \in \{10,20,30,50\}$, $\beta \in \{0.1,0.2,0.3\}$
  \\ \midrule 
  SynPO-v4 & $-\sigma\left(\alpha \cdot \overline{S(y_w)}-\alpha \cdot \overline{S(y_l)}\right)-\beta \, \overline{\log S(y_w)}$ 
  & $\alpha \in \{10,20,30,50\}$, $\beta \in \{0.05,0.1,0.2,0.3\}$
  \\  \midrule 
  SynPO-v5 &  $-\left(\alpha \cdot \exp \left(\overline{\log S(y_w)}\right)-\alpha \cdot \exp \left(\overline{\log S(y_l)}\right)\right)-\beta \, \overline{S(y_w)}$ 
  & $\alpha = 1$, $\beta \in \{0.01,0.02,0.05,0.1,0.15,0.2\}$
  \\ 
  \bottomrule
  \end{tabular}}
\end{table*}

\section{Implementation Details for Video Captioning}\label{sec:xperimental setup}
In Table \ref{tab:workflow-comparison}, we conduct ablation studies to evaluate the effectiveness of our data generation approach and SynPO. In the first experiment, we fine-tune AuroraCap using DPO and its various variants. The fine-tuning dataset is generated using our proposed data generation pipeline on a subset of ShareGPT4Video. In the second part of Table \ref{tab:workflow-comparison}, we only change the source dataset used in the data generation pipeline to verify the general adaptability of our pipeline and SynPO. In the third part of Table \ref{tab:workflow-comparison}, we use the same preference pairs generated from ShareGPT4Video as in the first experiment to fine-tune several popular multimodal models, and evaluate the resulting models.

\textbf{Data Generation Setup.} 
For the video detailed captioning experiment evaluating the effectiveness of our approach (Table \ref{table:greedy}), we sample over 10,000 source videos from the ShareGPT4Video dataset. We employ a sampling strategy with a count of 10 per video, using temperature = 0.9, top\_p = 0.95, and top\_k = 32 to generate diverse candidate captions. The LLM used to score all generated captions is Qwen-Plus-2025-01-25. Among the candidates, the caption receiving the highest score is selected as the positive preference, while the one with the lowest score is treated as the negative preference.

\textbf{Training Setup.} 
The maximum number of training epochs is set to 5, and the best-performing model based on validation performance is selected for final evaluation. We use the AdamW optimizer with a linear learning rate scheduler incorporating warmup. The warmup ratio is set to 0.1, and the learning rate is $5\times 10^{-6}$. The batch size is fixed at 32 during training. Our fine-tuning procedure follows a LoRA-based parameter-efficient configuration, which includes the following hyperparameters:

\begin{itemize}
\item \textbf{Rank:} Set to 128, controlling the dimensionality of the low-rank matrices used for adaptation.

\item \textbf{Lora\_alpha:} Set to 64, scaling the magnitude of the low-rank updates during training.

\item \textbf{Dropout:} Set to 0.05, introducing regularization by randomly zeroing out 5\% of the activations in the adapted layers.

\item \textbf{Target modules:} All linear projection layers are targeted for adaptation.

\end{itemize}

\label{intro to VDD and VDC}
\textbf{Evaluation Setup.}
During inference, we adopt a greedy decoding strategy for caption generation. Additionally, four widely-used benchmark datasets are employed to comprehensively evaluate the model's performance across multiple dimensions:
\begin{itemize}

\item Video Detailed Captioning (VDC)~\cite{chai2024auroracap} transforms the matching between two paragraphs into a set of question-answer pairings. It first generates some question-answer pairs based on the ground truth captions, then derive corresponding answers one by one from the generated captions, and finally perform matching. The process is automatically evaluated with the LLM involvement in each step.

\item Video Detailed Description (VDD)~\cite{li2023videochat} is a multimodal benchmark designed to evaluate models' ability to generate temporally coherent, semantically rich, and contextually precise natural language descriptions of video content, integrating visual, and textual modalities through datasets with fine-grained captions to challenge cross-modal reasoning, dynamic scene understanding, and long-term temporal modeling in video-language tasks. Notably, it utilizes LLM to score the similarity between ground-truth caption and generated caption.

\item Microsoft Research Video to Text (MSRVTT)~\cite{xu2016msr} is a large-scale multimodal benchmark designed to evaluate models' ability to generate temporally coherent and contextually rich textual descriptions of video content, comprising 10,000 video clips annotated with 20 English sentences each via crowdsourcing, and featuring standardized train/validation/test splits across 20 diverse categories to challenge cross-modal reasoning and dynamic scene understanding in video-language tasks.

\item VATEX~\cite{wang2020vatexlargescalehighqualitymultilingual} is a large-scale multilingual multimodal benchmark designed for video captioning and cross-lingual machine translation tasks, comprising 41,250 video clips annotated with 825,000 English-Chinese subtitles (206,000 aligned pairs), emphasizing cross-modal reasoning, temporal coherence, and linguistic diversity to evaluate models' ability to generate context-aware descriptions and leverage visual-spatial cues for accurate multilingual translation.
\end{itemize}

\textbf{Computing Resources.}
All the training experiments in this paper were conducted on 4 $\times$ NVIDIA H800 (80G) GPUs.

\section{Details of Prompts}\label{sec:full prompts}
\subsection{Prompts Used in Data Construction Pipeline}
In our pipeline of preference pair generation, we employed three set of prompts for LLM to score generated caption. Prompts are given in the format of Python code.

\subsubsection{First Criterion}

\begin{table*}[h!]\centering

\begin{minipage}{0.99\columnwidth}\vspace{0mm}    \centering
\begin{tcolorbox} 
    \centering
    \small
     \hspace{-6mm}
    \begin{tabular}{p{0.99\columnwidth}}

\begin{minipage}{0.99\columnwidth}\vspace{0mm}

\VarSty{messages} = [
    \{
    
    "role":"system", 
    
    "content": "You are a helpful assistant."\}

    \{
    
    "role":"user", 
    
    "content": f"""
    
    I am going to provide you with several video caption 
captions generated by a multimodal model. The Caption 1 is a caption of the entire video, which needs to be evaluated. The subsequent captions are captions generated after I divided the long video into segments. I would like you to score the Caption 1 based on the captions of the subsequent segments. Note that the captions of the subsequent segments is not absolutely accurate, so please tolerate some minor deviations. The scoring range is an integer from 0 to 5, with the main evaluation metric being whether there are inconsistencies between the entities or actions mentioned in the first caption and those in the following captions (i.e., whether hallucinations occur). The higher the hallucination, the lower the score.

Caption 1 (what you need to evaluate): 

\{\VarSty{sample['caption1']}\}

Caption 2 (what you need to refer to it): 

\{\VarSty{sample['caption2']}\}

Caption 3 (what you need to refer to it): 

\{\VarSty{sample['caption3']}\}

\VarSty{...}

Respond in JSON format, for example: \{'reasoning': your reasoning, 'score': an integer\}

    """\}

]

\end{minipage}
    \end{tabular}
\end{tcolorbox}
    
\end{minipage}
\end{table*}
\quad
\quad
\newpage

\subsubsection{Second Criterion}

\begin{table*}[h]\centering
    \centering
    \begin{minipage}{0.99\columnwidth}
        \vspace{0mm}
        \begin{tcolorbox}
            \centering
            \small
            \hspace{-7mm}
            \begin{tabular}{p{0.99\columnwidth}}
                \begin{minipage}{0.99\columnwidth}
                    \vspace{0mm}
                    
                    \VarSty{messages} = [
                        \{
                        
                        "role":"system", 
                        
                        "content": "You are a helpful assistant."  
                        \}

                        \{
                        
                        "role":"user", 
                        
                        "content": f"""
                        
                            I am going to provide you with a video caption generated by a multimodal model. 
                            I would like you to score it on a scale from 0 to 5, with 5 being the highest score. 
                            The main criteria for scoring are: 

                            1. Whether the caption meets the requirements of the corresponding prompt. 
                               Prompt: \{\VarSty{sample ['prompt']}\}

                            2. Whether the caption is natural and coherent, 
                               using language appropriate for describing a video. 
                               For instance, if phrases like 'this image is...' are used, a lower score should be given.

                            3. If there is subjective evaluation in the caption, please lower some marks. 
                               If there is some objective inference in the caption, 
                               this does not affect the score.  

                            The caption: 

                            \{\VarSty{sample['caption']}\}

                            DO NOT PROVIDE ANY OTHER OUTPUT TEXT OR EXPLANATION. 
                            Respond in JSON format, for example: \{'score': 4\}
                            
                        """
                        \} 
                        
                    ]
                \end{minipage}
            \end{tabular}
        \end{tcolorbox}
    \end{minipage}
\end{table*}
\quad

\subsubsection{Third Criterion}

\begin{table*}[h]
    \centering
    \begin{minipage}{0.99\columnwidth}
        \vspace{0mm}
        \begin{tcolorbox}
            \centering
            \small
            \hspace{-6mm}
            \begin{tabular}{p{0.99\columnwidth}}
                \begin{minipage}{0.99\columnwidth}
                    \vspace{0mm}
                    
                    \VarSty{messages} = [
                        \{
                        
                        "role":"system", 
                        
                        "content": "You are a helpful assistant."\}

                \end{minipage}
            \end{tabular}
        \end{tcolorbox}
    \end{minipage}
\end{table*}
\begin{table*}[h]
    \centering
    \begin{minipage}{0.99\columnwidth}
        \vspace{0mm}
        \begin{tcolorbox}
            \centering
            \small
            \hspace{-6mm}
            \begin{tabular}{p{0.99\columnwidth}}
                \begin{minipage}{0.99\columnwidth}
                    \vspace{0mm}
                                            \{
                        
                        "role":"user", 
                        
                        "content": f"""
                        
                            I will provide you with 
                            \{\VarSty{sample['length']}\} 
                            captions of the same video generated by a multimodal model. 
                            I would like you to score these captions based on the model's self-consistency. 
                            In other words, give higher scores to captions 
                            that are semantically similar, and lower scores to captions 
                            that differ significantly from the others. 
                            The scoring range is integers from 0 to 3, with 3 being the highest score. 

                            The caption 1:
                            \{\VarSty{sample['caption1']}\}

                            The caption 2:
                            \{\VarSty{sample['caption2']}\}

                            \VarSty{...}
                            
                            Respond in JSON format, for example: 

                            \{
                                'reasoning': your reasoning,
                                'the score of caption 1': an integer,
                                'the score of caption 2': an integer,
                                \VarSty{...}
                            \}
                            
                        """\}  
                        
                    ]
                \end{minipage}
            \end{tabular}
        \end{tcolorbox}
    \end{minipage}
\end{table*}

\subsection{Prompts Used in Token Experiments}

For our experiment in the figure in the main text, a set of prompts is required to score the semantic importance of a token. Detailed prompts are as follows:
\begin{table*}[h]
    \centering
    \begin{minipage}{0.99\columnwidth}
        \vspace{0mm}
        \begin{tcolorbox}
            \centering
            \small
            \hspace{-6mm}
            \begin{tabular}{p{0.99\columnwidth}}
                \begin{minipage}{0.99\columnwidth}
                    \vspace{0mm}
                    
                    \VarSty{messages} = [
                        \{
                        
                        "role":"system",
                        
                        "content": "You are a helpful assistant."\}
                        
                       \{
                        
                        "role":"user",
                        
                        "content": f"""

                            I am now preparing to analyze the importance 
                            of each token in a given sentence. I hope you can score the tokens I provide 
                            based on their significance in determining the semantic direction of the sentence. 
                            Note that some words are split into subwords, and in such cases, 
                            the first subword is more important than the subsequent ones. 
                            The scoring range is an integer between 0 and 5, with 5 being the highest score.

                            Sentence: 
                            \{\VarSty{sample['sentence']}\}

                            Token segmentation: 
                            \{\VarSty{sample['token\_segmentation']}\}

                            Do not output explanations. Only provide the results in JSON format with index numbers. 

                            Example: 

                            \{
                                "0": \{"The": 1\},
                                "1": \{"video": 4\},
                                ...
                            \}
                            
                        """\}    
                        
                    ]
                \end{minipage}
            \end{tabular}
        \end{tcolorbox}
    \end{minipage}
\end{table*}

\subsection{Prompts Used in Enhanced Inference Evaluation}
For our experiments in the figure in the main text, two sets of prompts are used to evaluate the inference results in six dimensions.

\subsubsection{Accuracy, Richness, Completeness and Fluency}
\quad
\begin{table*}[h]
    \centering
    \begin{minipage}{0.99\columnwidth}
        \vspace{0mm}
        \begin{tcolorbox}
            \centering
            \small
            \hspace{-6mm}
            \begin{tabular}{p{0.99\columnwidth}}
                \begin{minipage}{0.99\columnwidth}
                    \vspace{0mm}
                    
                    \VarSty{messages} = [
                        \{
                        
                        "role":"system", 
                        
                        "content": "You are a helpful assistant."\}

                \end{minipage}
            \end{tabular}
        \end{tcolorbox}
    \end{minipage}
\end{table*}

\begin{table*}[h]
    \centering
    \begin{minipage}{0.99\columnwidth}
        \vspace{0mm}
        \begin{tcolorbox}
            \centering
            \small
            \hspace{-6mm}
            \begin{tabular}{p{0.99\columnwidth}}
                \begin{minipage}{0.99\columnwidth}
                        \{
                        
                        "role":"user", 
                        
                        "content": f"""
                        
                            I will provide you with two captions of a video. 
                            The first caption is correct, and the second one is generated by a multimodal model. 
                            I would like you to evaluate the second caption 
                            based on four different criteria, each scored as an integer between 0 and 5, 
                            where 5 is the highest score. 
                            The scoring criteria are as follows:

                            1. How many inaccurate or fabricated details are present in the second caption; 
                               the more inaccuracies, the lower the score.

                            2. How rich in detail the second caption is; 
                               the more details, the higher the score.

                            3. How well the second caption captures the main elements of the video.

                            4. Whether the second caption matches the tone and style expected 
                               for describing a video, and whether the sentences are fluent and natural.

                            The correct caption: \{\VarSty{sample['answer']}\}

                            The predicted caption: \{\VarSty{sample['pred']}\}

                            Respond in JSON format, for example: 

                            \{
                                "analysis": "Your evaluation process and scoring rationale",
                                "score 1": "An integer",
                                "score 2": "An integer",
                                "score 3": "An integer", 
                                "score 4": "An integer"
                            \}
                            
                        """\}    
                        
                    ]

                 \end{minipage}
            \end{tabular}
        \end{tcolorbox}
    \end{minipage}
\end{table*}

\subsubsection{Dynamics and Coherence}
\quad
\begin{table*}[h]
    \centering
    \begin{minipage}{0.99\columnwidth}
        \vspace{0mm}
        \begin{tcolorbox}
            \centering
            \small
            \hspace{-6mm}
            \begin{tabular}{p{0.99\columnwidth}}
                \begin{minipage}{0.99\columnwidth}
                    \vspace{0mm}
                    
                    \VarSty{messages} = [
                        \{
                        
                        "role":"system", 
                        
                        "content": "You are a helpful assistant."\}
                        
                        \{
                        
                        "role":"user", 
                        
                        "content": f"""
                        
                            I will provide you with two captions of a video. 
                            The first caption is correct, and the second one is generated by a multimodal model. 
                            I would like you to evaluate the second caption 
                            based on two different criteria, each scored as an integer between 0 and 5, 
                            where 5 is the highest score. 
                            The scoring criteria are as follows:

                            1. How accurately the second caption captures temporal changes, 
                               such as possible actions of people or animals, or shifts in the scene.

                            2. Whether the development of events in the second caption is coherent and consistent, 
                               following a logical time sequence.

                            The correct caption: \{\VarSty{sample['answer']}\}

                            The predicted caption: \{\VarSty{sample['pred']}\}

                            Respond in JSON format, for example:

                            \{
                                "analysis": "Your evaluation process and scoring rationale",
                                "score 1": "An integer", 
                                "score 2": "An integer"
                            \}
                            
                        """\}   
                        
                    ]
                \end{minipage}
            \end{tabular}
        \end{tcolorbox}
    \end{minipage}
\end{table*}

\section{Mathematical Derivation}
\subsection{Deriving the Objective function of DPO}
For RLHF, the first step is to train the reward model. The training data consists of two responses to the same prompt, where human annotators or GPT-4 label which response is better. The reward model optimizes the following loss:  
$$
\max_{r_{\phi}}\left\{\mathbb{E}_{(x,y_\text{win},y_\text{lose})\sim\mathcal{D}}[\log\sigma(r_\phi(x,y_\text{win})-r_\phi(x,y_\text{lose}))]\right\}
$$  
Here, $ r_\phi $ is the reward model used to score responses, $ \mathcal{D} $ denotes the training dataset, $ x $ is the prompt, and $ y_\text{win} $ and $ y_\text{lose} $ represent the better and worse responses, respectively. This formulation aims to maximize the score difference between better and worse responses.  

The second step employs an RL algorithm to improve the model's scores. The loss function is defined as:  
$$
\max_{\pi_\theta}\left\{\mathbb{E}_{x\sim \mathcal{D},y\sim\pi_\theta(y|x)}[r_\phi(x,y)]-\beta\mathbb{D}_{\text{KL}}[\pi_\theta(y|x)||\pi_\text{ref}(y|x)]\right\}
$$  
where $ \pi_\theta $ represents the LLM being trained, and $ \pi_\text{ref} $ is the initial reference model. This loss function aims to maximize the reward scores of the LLM's outputs while ensuring $ \pi_\theta $ does not deviate excessively from $ \pi_\text{ref} $, maintaining the model's ability to generate coherent responses rather than producing high scores but nonsensical outputs.  

The authors of DPO recognized that the latter expression admits an explicit solution. Specifically:  
$$
\begin{aligned}
\max_{\pi_\theta}&\left\{\mathbb{E}_{x\sim \mathcal{D},y\sim\pi_\theta(y|x)}[r_\phi(x,y)] -\beta\mathbb{D}_{\text{KL}}[\pi_\theta(y|x)||\pi_\text{ref}(y|x)]\right\}\\
&=\max_{\pi_\theta}\mathbb{E}_{x\sim \mathcal{D},y\sim\pi_\theta(y|x)}[r_\phi(x,y) - \beta \log \frac{\pi_\theta(y|x)}{\pi_\text{ref}(y|x)}]\\
&=\min_{\pi_\theta}\mathbb{E}_{x\sim \mathcal{D},y\sim\pi_\theta(y|x)}[\log \frac{\pi_\theta(y|x)}{\pi_\text{ref}(y|x)} - \frac{1}{\beta} r_\phi(x,y)]\\
&=\min_{\pi_\theta}\mathbb{E}_{x\sim \mathcal{D},y\sim\pi_\theta(y|x)}[\log\frac{\pi_\theta(y|x)}{\pi_\text{ref}(y|x)e^{r_\phi(x,y)/\beta}}]
\end{aligned}
$$  

By normalizing the denominator (i.e., setting $ Z(x)=\sum_y\pi_\text{ref}(y|x)e^{r_\phi(x,y)/\beta} $), we can construct a new probability distribution:  
$$
\pi^*(y|x) = \pi_\text{ref}(y|x)e^{r_\phi(x,y)/\beta}/Z(x)
$$  
Substituting this into the previous expression yields:  
$$
\begin{aligned}
\min_{\pi_\theta}&\mathbb{E}_{x\sim \mathcal{D},y\sim\pi_\theta(y|x)}[\log\frac{\pi_\theta(y|x)}{\pi_\text{ref}(y|x)e^{r_\phi(x,y)/\beta}}]\\ 
&=\min_{\pi_\theta}\mathbb{E}_{x\sim \mathcal{D},y\sim\pi_\theta(y|x)}[\log\frac{\pi_\theta(y|x)}{\pi^*(y|x)}-\log Z(x)]\\ 
&=\min_{\pi_\theta}\mathbb{E}_{x\sim \mathcal{D},y\sim\pi_\theta(y|x)}[\log\frac{\pi_\theta(y|x)}{\pi^*(y|x)}]\\ 
&=\min_{\pi_\theta}\mathbb{E}_{x\sim \mathcal{D}}\mathbb{D}_\text{KL}(\pi_\theta(y|x)||\pi^*(y|x)) 
\end{aligned}
$$  
Since the KL divergence achieves its minimum when the two distributions are equal, we conclude that the optimal probability distribution under RLHF training is $ \pi^* $.  

Alternatively, from the definition of $ \pi^* $, we derive a relationship between $ r_\phi $ and $ \pi^* $.  We can directly train $ \pi^* $ instead of $ r_\phi $. By rearranging the definition of $ \pi^* $, we obtain:  
$$
r_{\phi}(x,y)=\beta\log\frac{\pi^*(y|x)}{\pi_\text{ref}(y|x)}+\beta \log Z(x)
$$  
Substituting this into the original loss for optimizing $ r_\phi $ leads to:  
$$
\max_{\pi^*}\left\{\mathbb{E}_{(x,y_\text{win},y_\text{lose})\sim\mathcal{D}}[\log\sigma(\beta\log\frac{\pi^*(y_\text{win}|x)}{\pi_\text{ref}(y_\text{win}|x)} - \beta\log\frac{\pi^*(y_\text{lose}|x)}{\pi_\text{ref}(y_\text{lose}|x)})]\right\}
$$  
Equivalently, we can directly optimize $ \pi_\theta $ using this loss:  
$$
\max_{\pi_\theta}\left\{\mathbb{E}_{(x,y_\text{win},y_\text{lose})\sim\mathcal{D}}[\log\sigma(\beta\log\frac{\pi_\theta(y_\text{win}|x)}{\pi_\text{ref}(y_\text{win}|x)} - \beta\log\frac{\pi_\theta(y_\text{lose}|x)}{\pi_\text{ref}(y_\text{lose}|x)})]\right\}
$$  
This is the DPO loss. By transforming the above equations, DPO smoothly converts RLHF into SFT. During training, it no longer requires running four models simultaneously (reward model, ref model, critic, and actor), but only two models (actor and ref). Furthermore, since online data sampling is no longer required, the outputs of the ref model can be precomputed and reused during training.

\subsection{Deriving the Gradient of the DPO Objective}
\label{app:gradient_derivation}
In this section we derive the gradient of the DPO objective:
\begin{align}\label{eq:grad-start}
    \nabla_{\theta}\mathcal{L}_\text{DPO}(\pi_{\theta}; \pi_{\mathrm{ref}})
    = -\nabla_{\theta}\mathbb{E}_{(x, y_w, y_l)\sim \mathcal{D}}\left[\log \sigma \left(\beta \log \frac{\pi_{\theta}(y_l|x)}{\pi_{\mathrm{ref}}(y_l|x)} - \beta \log \frac{\pi_{\theta}(y_w|x)}{\pi_{\mathrm{ref}}(y_w|x)}\right)\right]
\end{align}
We can rewrite the RHS of Equation~\ref{eq:grad-start} as 
\begin{align}
    \nabla_{\theta}\mathcal{L}_\text{DPO}(\pi_{\theta}; \pi_{\mathrm{ref}})
    =-\mathbb{E}_{(x, y_w, y_l)\sim \mathcal{D}}\left[\frac{\sigma'\left(u\right)}{\sigma \left(u\right)}\nabla_{\theta}\left(u\right)\right],
\end{align}
where $u = \beta \log \frac{\pi_{\theta}(y_l|x)}{\pi_{\mathrm{ref}}(y_l|x)} - \beta \log \frac{\pi_{\theta}(y_w|x)}{\pi_{\mathrm{ref}}(y_w|x)}$.
Using the properties of sigmoid function $\sigma'(x) = \sigma(x)(1-\sigma(x))$ and $\sigma(-x) = 1-\sigma(x)$, we obtain the final gradient
\begin{multline*}
\nabla_{\theta}\mathcal{L}_\text{DPO}(\pi_{\theta}; \pi_{\mathrm{ref}}) = \\
     -\mathbb{E}_{(x, y_w, y_l) \sim \mathcal{D}} \bigg[\beta\sigma \left(\beta \log \frac{\pi_{\theta}(y_w|x)}{\pi_{\mathrm{ref}}(y_w|x)} - \beta \log \frac{\pi_{\theta}(y_l|x)}{\pi_{\mathrm{ref}}(y_l|x)}\right)\bigg[\nabla_\theta\log \pi_{\theta}(y_w \mid x) - \nabla_\theta\log\pi_{\theta}(y_l \mid x)\bigg]\bigg],
\end{multline*}
After using the reward substitution of $\hat{r}_\theta(x, y) = \beta \log \frac{\pi_\theta(y \mid x)}{\pi_{\mathrm{ref}}(y \mid x)}$ we obtain the final form of the gradient.

\subsection{Deriving the Gradient of DPO with respect to reward function}
In this section we derive the gradient demonstrated in the figure in the main text (about the normalized Frobenius norm of the gradient). Note that for original DPO, 
$$
r(x,y) =\beta \log \frac{\pi_r(y\mid x)}{\pi_{\mathrm{ref}}(y\mid x)} + \beta \log Z(x).$$
thus we have:
\begin{align*}
    \dfrac{\partial \mathcal{L}_{DPO}}{\partial r_+} &= \dfrac{\partial\log \sigma (r_+-r_-)}{\partial r_+} \\
    &= \dfrac{\partial\log \sigma (r_+-r_-)}{\partial \sigma (r_+-r_-)} \cdot \dfrac{\partial \sigma (r_+-r_-)}{\partial(r_+-r_-)}\cdot \dfrac{\partial(r_+-r_-)}{\partial r_+}\\
    &=\dfrac{1}{\sigma(r_+-r_-)}\cdot \sigma(r_+-r_-)[1-\sigma(r_+-r_-)]
\end{align*}
By analogy we have:
$$\dfrac{\partial \mathcal{L}_{DPO}}{\partial r_-}=-\dfrac{1}{\sigma(r_+-r_-)}\cdot \sigma(r_+-r_-)[1-\sigma(r_+-r_-)]$$

\section{Case Study}
We randomly chose a video from Panda70M dataset and compared captions generated by AuroraCap and our SynPO-fine-tuned version. We manually analyzed their differences, thus further demonstrate the effectiveness of SynPO. During inference, models are required to describe in detail what is happening in the video, including the subject matter, the setting, and possible character activities.
\subsection{Case 1}
Caption generated by AuroraCap:
\begin{verbatim}
    In the image, a man dressed as Batman is interacting with children at 
    a school cafeteria. The man is standing in the middle of the cafeteria, 
    surrounded by children who are seated at tables. He is wearing a Batman 
    costume, complete with a mask and a cape. The children are eating lunch, 
    and there are trays of food in front of them. The setting appears to be a 
    school cafeteria, as indicated by the presence of children and the food trays. 
    The man in the Batman costume is likely participating in a charitable event 
    or a special visit to the school, as suggested by the text overlay on the 
    image. The text overlay reads "Batman visits N.Y. school for cancer awareness." 
    This suggests that the man in the Batman costume is there to raise awareness 
    about cancer, possibly as part of a fundraising or awareness campaign. 
    The children seem to be enjoying the interaction with the man in the costume, 
    adding a fun and engaging element to the event.
\end{verbatim}

Caption generated by our fine-tuned version:
\begin{verbatim}
    The video clip captures a moment of joy and surprise at a school cafeteria 
    where a costumed character, resembling a superhero, is interacting with 
    children. The character, dressed in a black Batman-like suit, is standing 
    over a table filled with children's lunch boxes. The children, some of whom 
    are wearing glasses, are seated at the table, eating and engaging with the 
    character. The setting is a school cafeteria, indicated by the presence of 
    milk cartons and the children's casual attire. The character's costume and 
    the children's reactions suggest a special event, possibly a visit to raise 
    awareness or morale. The scene is lively, with the character's costume adding 
    a sense of excitement and wonder to the children's environment.
\end{verbatim}

Analysis: The second paragraph demonstrates superior descriptive accuracy in three key aspects: First, it avoids subjective speculation by refraining from definitive claims like "charitable event" and "fundraising" found in the first paragraph. Instead, it adopts neutral phrasing such as "special event" or "morale-boosting visit," which aligns with the video's inherent ambiguity. Second, the second paragraph enhances scene authenticity through meticulous detail selection—replacing the generic "food trays" with specific elements like "milk cartons" and "casual attire." This preserves the cafeteria setting while avoiding overgeneralization. Finally, in terms of linguistic expression, the second paragraph employs dynamic vocabulary such as "joy and surprise," "lively," and "wonder" to construct an emotional atmosphere. In contrast, the first paragraph, despite its detailed description of the Batman costume and text overlay, undermines the scene's vividness by fixating on specific elements like "cape" and "mask." This balanced combination of precise objective observation and effective emotional communication allows the second paragraph to more faithfully convey the video's core message.

\subsection{Case 2}
Caption generated by AuroraCap:
\begin{verbatim}
    The video captures a moment of a person walking on a snowy sidewalk. 
    The individual is dressed in winter attire, including a black coat 
    and boots, which suggests that the weather is cold. The person's posture 
    and the direction of their gaze indicate that they are walking away 
    from the camera, possibly towards a destination out of frame. The 
    presence of a vehicle in the background suggests that this might be a 
    residential area or a street with parked cars. The snow-covered ground 
    and the person's winter clothing imply that the scene is set during the 
    winter season. There are no visible signs of other activities or characters 
    in the immediate vicinity of the person. The overall setting appears to 
    be a quiet, possibly suburban, street during the daytime.
\end{verbatim}

Caption generated by our fine-tuned version:
\begin{verbatim}
    The video clip captures a moment of a person walking on a snow-covered 
    sidewalk. The individual, clad in a black coat and boots, is in mid-stride, 
    suggesting movement. The snow-covered ground and the white van in the 
    background indicate a winter setting. The person's attire, including the 
    boots and coat, is appropriate for the cold weather, and the snowy conditions. 
    The scene is quiet, with no other individuals or vehicles in immediate view, 
    which places the focus on the solitary figure. The person's posture and the 
    environment suggest a casual, possibly routine walk, perhaps to or from a 
    destination. The overall scene is a snapshot of a winter day, with the 
    individual's presence adding a sense of life and activity to the otherwise 
    still and quiet setting.
\end{verbatim}

Analysis: The second paragraph emphasizes motion and immediacy through phrases like "in mid-stride, suggesting movement" and "snapshot of a winter day," which capture the temporal flow of the video. In contrast, the first paragraph uses static descriptions ("walking away from the camera") and lacks nuance about the person’s actions. Video captioning benefits from dynamic language that reflects the unfolding of events rather than static snapshots. The second paragraph builds a cohesive narrative by linking the individual’s attire ("appropriate for cold weather") to the environmental context ("snow-covered ground") and infers a subtle emotional tone ("sense of life and activity in an otherwise still setting"). This creates a richer, more immersive description that mirrors how viewers naturally interpret visual scenes. The first paragraph, while factual, remains fragmented and lacks the emotional and contextual depth that enhances viewer engagement.

\subsection{Case 3}
Caption generated by AuroraCap:
\begin{verbatim}
    The image captures a moment from a weightlifting competition, with a 
    woman in the center of the frame performing a weightlifting exercise. 
    She is in the process of lifting a barbell, which is positioned above 
    her head. The woman is wearing a green and white outfit, which is 
    typical attire for weightlifting competitions.
    The setting appears to be a gym or a sports facility, as there are 
    several people watching the woman lift the barbell. Some of them are 
    seated on chairs, while others are standing. The audience is focused 
    on the woman's performance, indicating that this is an important event.
    The image also includes a clock on the wall, which is likely used to 
    keep track of the time during the competition. The presence of the 
    clock suggests that the woman's performance is being timed, and she 
    needs to lift the barbell within a specific time frame.
    Overall, the image conveys a sense of anticipation and focus, as the 
    woman prepares to lift the barbell, and the audience watches intently. 
    The setting and the woman's attire indicate that this is a formal 
    weightlifting competition, and the audience's attention suggests that 
    the woman's performance is significant.
\end{verbatim}

Caption generated by our fine-tuned version:
\begin{verbatim}
    The video captures a weightlifting event in an indoor gym, 
    where a female athlete is in the midst of a squat, her body poised 
    in a powerful stance. She is wearing a green and white singlet, 
    which is typical attire for weightlifting competitions. The barbell, 
    loaded with a substantial weight, is held high above her head, 
    indicating the peak of her lift. The audience, composed of spectators 
    and possibly coaches, is seated on chairs and benches, their attention 
    riveted on the athlete's performance. The gym is equipped with various 
    weightlifting machines and a mirrored wall, reflecting the ongoing 
    activity. The scene is a testament to the athlete's strength and the 
    supportive environment of the gym. The video provides a dynamic snapshot 
    of the sport, showcasing the athlete's form, the weight of the barbell, 
    and the engaged audience.
\end{verbatim}

Analysis: he second paragraph employs vivid, action-oriented language (e.g., "midst of a squat," "poised in a powerful stance," "dynamic snapshot") to capture the real-time motion and intensity of the weightlifting act. In contrast, the first paragraph relies on static descriptions ("performing a weightlifting exercise") and lacks temporal specificity, failing to convey the athlete’s movement progression or the urgency of the lift. Video captioning thrives on dynamic phrasing that mirrors the temporal flow of visual content. The second paragraph integrates environmental cues (e.g., "mirrored wall," "weightlifting machines") to enrich the scene’s authenticity, while the first paragraph underutilizes these elements. Additionally, the second paragraph subtly ties the athlete’s physicality ("powerful stance") to the gym’s functional design, creating a cohesive narrative that reflects the interplay between subject and environment. The first paragraph, though detailed, remains fragmented and lacks this holistic integration.

\end{document}